\pdfoutput=1

\documentclass[11pt]{article}

\usepackage[final]{acl}

\usepackage{latexsym}
\usepackage{wrapfig}

\usepackage{amsmath,amsfonts,bm}

\def\eqref#1{equation~\ref{#1}}

\def\1{\bm{1}}

\DeclareMathAlphabet{\mathsfit}{\encodingdefault}{\sfdefault}{m}{sl}
\SetMathAlphabet{\mathsfit}{bold}{\encodingdefault}{\sfdefault}{bx}{n}

\usepackage{times}
\usepackage{color}
\usepackage{bm}
\usepackage{tabularx}
\usepackage{tabularray}
\usepackage{dsfont}
\usepackage[T1]{fontenc}

\usepackage[utf8]{inputenc}

\usepackage{microtype}

\usepackage{inconsolata}

\usepackage{graphicx}
\usepackage{microtype}

\usepackage{hyperref}       %
\usepackage{inconsolata}
\usepackage{algorithm}
\usepackage{wrapfig}
\usepackage{algorithmic}
\usepackage{colortbl}
\usepackage{wrapfig}
\usepackage{xcolor}
\usepackage{inconsolata}
\usepackage{multirow}  
\usepackage{amssymb}   
\usepackage{amsmath}
\usepackage{color}
\usepackage{fontawesome5}
\usepackage{subfigure} 
\usepackage{makecell} 
\usepackage{booktabs} 
\usepackage{enumitem} 
\usepackage{multirow} 

\usepackage{booktabs} 
\usepackage{stfloats}
\usepackage{listings}
\usepackage{cite}
\usepackage{pifont}
\usepackage{tcolorbox}
\usepackage{soul}
\usepackage{enumitem}
\tcbuselibrary{skins, breakable, theorems}
\usepackage{wrapfig}
\usepackage{lscape}
\usepackage{url}

\definecolor{mypink}{rgb}{.99,.91,.95}
\definecolor{myyellow}{rgb}{.99,.94,.82}
\newcommand{\ours}[0]{EAGLET}
\newcommand{\thought}[1]{\textcolor{blue!70!black}{\textbf{Thought:} #1}}
\newcommand{\observation}[1]{\textcolor{brown!80!black}{\textbf{Observation:} #1}}
\newcommand{\action}[1]{\textcolor{green!60!black}{\textbf{Action:} #1}}

\definecolor{Gray}{gray}{0.9}
\definecolor{mygreen}{rgb}{0.0, 0.5, 0.0}
\definecolor{myred}{rgb}{0.8, 0.25, 0.33}
\definecolor{myblue}{rgb}{0.19, 0.55, 0.91}
\definecolor{uclablue}{rgb}{0.15, 0.45, 0.68}
\definecolor{ucladblue}{rgb}{0.0, 0.33, 0.53}
\definecolor{ucladdblue}{rgb}{0.0, 0.23, 0.36}
\definecolor{uclagold}{rgb}{1.0, 0.82, 0.0}
\definecolor{ucladgold}{rgb}{1.0, 0.78, 0.17}
\definecolor{ucladdgold}{rgb}{1.0, 0.72, 0.11}
\definecolor{boxgreen}{rgb}{0.02, 0.66, 0.02}
\definecolor{boxred}{rgb}{0.66, 0.1, 0.1}
\definecolor{boxblue}{rgb}{0.01, 0.01, 0.73}

\usepackage{times}
\usepackage{latexsym}

\usepackage[utf8]{inputenc}
\usepackage[T1]{fontenc}

\usepackage{inconsolata}

\usepackage{algorithm}
\usepackage{algorithmic}

\usepackage{newfloat}
\usepackage{listings}
\floatstyle{ruled}
\newfloat{listing}{tb}{lst}{}
\floatname{listing}{Listing}

\usepackage{graphicx}
\usepackage{amsmath,amssymb,amsthm,mathabx,amsfonts}
\usepackage{bbm}
\usepackage{cleveref}
\usepackage{acronym}
\usepackage{enumitem}
\usepackage{balance}
\usepackage{xspace}
\usepackage{setspace}
\usepackage{url}
\usepackage{amsfonts}
\usepackage{multirow}
\usepackage{booktabs}
\usepackage{tablefootnote}
\usepackage[switch]{lineno}
\usepackage{multicol,lipsum}
\usepackage{tikz}
\usepackage{pgfplots}
\usepackage{pgfplotstable}
\usepackage{arydshln}

\usepackage{CJKutf8}

\pgfplotsset{compat=1.18}
\makeatletter
\DeclareRobustCommand\onedot{\futurelet\@let@token\@onedot}
\def\@onedot{\ifx\@let@token.\else.\null\fi\xspace}

\makeatother

\makeatletter
\newcommand{\thickhline}{%
    \noalign {\ifnum 0=`}\fi \hrule height 1pt
    \futurelet \reserved@a \@xhline
}

\crefname{algorithm}{Alg.}{Algs.}
\Crefname{algocf}{Algorithm}{Algorithms}
\crefname{section}{Sec.}{Secs.}
\Crefname{section}{Section}{Sections}
\crefname{table}{Tab.}{Tabs.}
\Crefname{table}{Table}{Tables}
\crefname{figure}{Fig.}{Fig.}
\Crefname{figure}{Figure}{Figure}
\crefname{appendix}{Appendix}{Appendices}

\acrodef{nlp}[NLP]{natural language processing}
\acrodef{plm}[PLM]{Pre-trained Language Model}
\acrodef{sota}[SOTA]{state-of-the-art}
\acrodef{icl}[ICL]{In-Context Learning}
\acrodef{bbl}[BBL]{BIG-bench Lite}

\usepackage{colortbl}
\definecolor{gblue}{HTML}{4285F4}
\definecolor{gred}{HTML}{DB4437}
\definecolor{ggreen}{HTML}{0F9D58}

\definecolor{mygray}{gray}{.92}
\definecolor{emphypurple}{rgb}{0.302, 0.055, 0.659}

\definecolor{highlightgreen}{HTML}{009901}
\definecolor{highlightred}{HTML}{FD6864}

\title{A Goal Without a Plan Is Just a Wish: Efficient and Effective Global Planner Training for Long-Horizon Agent Tasks}
\author{
\textbf{Shuzheng Si\thanks{\ Equal Contribution.}$^{\spadesuit\diamondsuit}$, Haozhe Zhao\footnotemark[1]$^{\clubsuit}$,  Kangyang Luo$^{\spadesuit}$, Gang Chen$^{\diamondsuit}$} \\ 
\textbf{Fanchao Qi\thanks{\ Corresponding Authors.}$^{\diamondsuit}$, Minjia Zhang$^{\clubsuit}$, Baobao Chang$^{\heartsuit}$,} and \textbf{Maosong Sun\footnotemark[2]$^{\spadesuit}$} \\ 
$^{\spadesuit}$ Tsinghua University \quad $^{\heartsuit}$ Peking University
\quad $^\diamondsuit$ DeepLang AI \\
\quad $^{\clubsuit}$ University of Illinois Urbana-Champaign
\\
\texttt{ssz24@mails.tsinghua.edu.cn}
}

\begin{document}

\maketitle

\renewcommand{\thefootnote}{\fnsymbol{footnote}}
\renewcommand{\thefootnote}{\arabic{footnote}}
\urlstyle{same}
\definecolor{darkgreen}{RGB}{50,100,0}
\definecolor{darkred}{RGB}{200, 0, 0}
\definecolor{lightred}{RGB}{250, 200, 200}
\definecolor{lightblue}{RGB}{210, 220, 250}
\newcommand{\cmark}{\textcolor{darkgreen}{\ding{51}}} %
\newcommand{\xmark}{\textcolor{darkred}{\ding{55}}} %
\definecolor{tabcolor1}{RGB}{247,225, 237} 
\definecolor{tabcolor2}{RGB}{255, 250, 132} 
\definecolor{tabcolor3}{RGB}{204, 232, 207} 
\definecolor{tabcolor4}{RGB}{245, 222, 179} 
\definecolor{tabcolor5}{RGB}{210, 220, 250} 
\definecolor{tabcolor6}{RGB}{222, 222, 222} 

\begin{abstract} 
Agents based on large language models (LLMs) struggle with brainless trial-and-error and generating hallucinatory actions due to a lack of global planning in long-horizon tasks.
In this paper, we introduce a plan-and-execute framework and propose \textbf{EAGLET}, an efficient and effective planner training method to enhance the executor agent's planning abilities without human effort.
Specifically, we train a plug-and-play global planner through a two-step process: we first synthesize high-quality plans from an advanced LLM using our proposed \textit{homologous consensus filtering} strategy, and apply fine-tuning as a cold start. 
Moreover, we further improve the planner with a rule-based reinforcement learning stage using a novel \textit{executor capability gain reward}, ensuring it can handle task instructions of varying difficulty.
Experiments on three long-horizon agent tasks show that executor agents equipped with our planner outperform existing methods, achieving new state-of-the-art performance. 
Meanwhile, EAGLET reduces training costs by 8× compared to RL-based baselines, and it does not require manual effort or extra training data, offering an efficient and effective solution.
\end{abstract}
\section{Introduction}
\label{section:introduction}
Recent progress in large language models (LLMs) \citep{OpenAI2023GPT4TR, Claude3S} has leapt from static chatbots to versatile agents that tackle complex tasks, such as science experiments \citep{kon2025curierigorousautomatedscientific}.
For these long-horizon tasks \citep{zhang2024surveymemorymechanismlarge}, agents need to handle multi-turn interactions, reason about actions, and adapt to dynamic environments \citep{si2023spokenwoz, huang2024understandingplanningllmagents}.

However, as LLMs are autoregressive models trained with next-token prediction, they often lack the planning ability needed for long-horizon agent tasks \citep{feng2025groupingrouppolicyoptimizationllm}, leading to performing brainless trial and error in the environment \citep{qiao2024agent,zhu-etal-2025-knowagent}.
When leveraging the model’s inner ability and performing on-the-fly planning during task execution, e.g., ReAct \citep{yao2023react}, LLM-based agents with static prompt engineering are prone to generating hallucinatory actions due to \textit{planning hallucinations} \citep{zhu-etal-2025-knowagent,xiong2025mpoboostingllmagents}.
This raises a critical question: \textbf{\textit{how can we improve the planning abilities of LLM-based agents to mitigate planning hallucinations for long-horizon agent tasks?}}

\begin{figure}
    \centering
    \includegraphics[width=7.7cm]{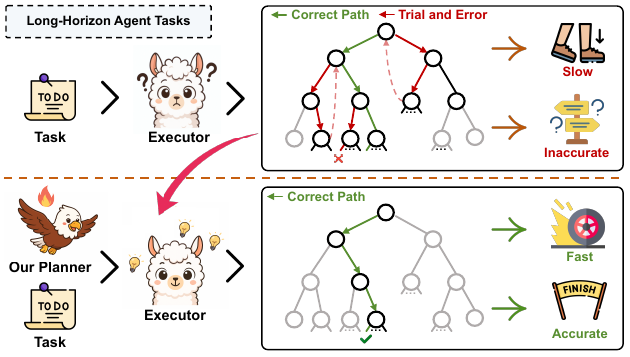}
    \caption{Traditional agent planning vs. Agent planning with our planner EAGLET.
    In this way, executor agents can complete tasks better within fewer interactions.}
    \label{fig_example}
\end{figure}

The first line of research attempts to apply supervised fine-tuning (SFT) or reinforcement learning (RL) so agents can implicitly learn planning abilities from expert-annotated trajectories and agent-environment interactions \citep{yin-etal-2024-agent, fei2025unleashingembodiedtaskplanning}.
For example, \citet{zeng-etal-2024-agenttuning} develops LLM-based agents by implementing SFT on collected expert trajectories.
Inspired by outcome-based RL methods \citep{shao2024deepseekmathpushinglimitsmathematical}, \citet{feng2025groupingrouppolicyoptimizationllm} proposes a novel RL algorithm GiGPO that achieves fine-grained reward assignment for enhancing agents.
However, these \textbf{\textit{implicit planning methods}} regard the agent as just an executor and focus on local planning, where planning only occurs through interleaved reasoning and action generation \citep{wang2025policyoptimizationdatacuration}.
This makes these methods less effective for long-horizon tasks requiring multi-step execution and global planning.
Also, implicit planning methods are inefficient: SFT-based methods are data-inefficient, requiring substantial amounts of expert-annotated data to generalize well due to deficient exploration of environments; 
RL-based methods are training-inefficient, requiring extensive training time and iterations to converge because rewards are delayed and sparse, and episodes often have tens of decision steps.




\begin{figure*}
    \centering
    \includegraphics[width=1\linewidth]{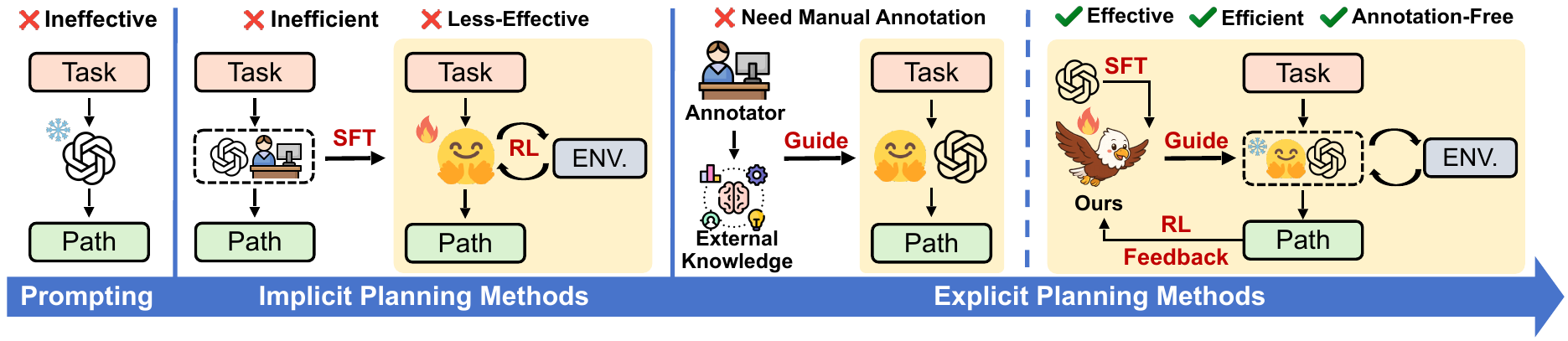}
    \caption{EAGLET vs. previous methods: we introduce a plug-and-play, efficient, and effective global planner to provide explicit guidance to mitigate planning hallucinations without human effort.
    }
    \label{figure:intro_2}
\end{figure*}

Beyond these implicit planning methods, \textbf{\textit{explicit planning methods}} try to use explicit knowledge to guide agents in task execution \citep{qiao2024agent, xiong2025mpoboostingllmagents}.
For example, KnowAgent \citep{zhu-etal-2025-knowagent} builds an action knowledge base to explicitly support global planning, enabling more reasonable trajectories.
MPO \citep{xiong2025mpoboostingllmagents} proposes training a global planner to provide high-level guidance that assists in agent planning.
Even though effective, these methods often need substantial manual efforts, such as manual verification and modification of collected data \citep{xiong2025mpoboostingllmagents}, making them hard and inefficient to transfer across different tasks and environments.

To tackle these challenges in both efficiency and effectiveness, we propose an \underline{e}fficient \underline{a}nd effective \underline{g}lobal p\underline{l}ann\underline{e}r \underline{t}raining method (\textbf{EAGLET}) to enhance the planning ability for long-horizon agent tasks without manual effort.
Inspired by \citet{wang-etal-2023-plan, qiao2024agent, xiong2025mpoboostingllmagents}, we introduce a plug-and-play and task-specific planner model whose sole responsibility is to generate a global plan shown in Figure \ref{fig_example}. 
Unlike implicit planning methods \citep{yin-etal-2024-agent, song-etal-2024-trial, feng2025groupingrouppolicyoptimizationllm} that regard the agent as just an executor and focus on local planning, we utilize a plan-and-execute framework where the global planner and executor are cleanly separated.
By decoupling high-level planning from local action execution, our framework explicitly endows the agent with the global foresight that implicit planning methods lack, therefore mitigating planning hallucinations.
Moreover, we introduce an efficient yet powerful pipeline for training a global planner without manual labor.
Specifically, we first query the advanced LLM to synthesize expert-level global plans, then we propose a novel \textit{homologous consensus filtering} method to ensure the quality of the synthesized plans. 
After applying the SFT stage on these filtered plans as a cold start for training a planner, we introduce a rule-based RL stage with a well-designed reward signal called \textit{executor capability gain reward (ECGR)} to enhance the generalization abilities of the global planner.
This reward thoroughly assesses a plan by checking if it can help multiple homologous executor agents with different capability levels to complete tasks better within fewer interactions. 
Finally, our efficient and labor-free training pipeline yields a plug-and-play planner model, which can provide expert-level and global foresight that facilitates task completion for any new executor agent shown in Figure \ref{figure:intro_2}.

We test our method on three long-horizon agent tasks: ScienceWorld \citep{wang-etal-2022-scienceworld}, ALFWorld \citep{shridhar2021alfworld}, and WebShop \citep{yao2022webshop}.
Executor agents equipped with our trained planner outperform those without it, achieving new state-of-the-art results.
Also, our proposed EAGLET can reduce training time by 8× compared with applying RL-based methods on executor agents, while eliminating the need for any human effort and additional training data.










\section{Task Formulation}
\label{section:method}

Our work focuses on improving the global planning of LLM-based agents for task completion. 
During the agent-environment interactions, the executor agent’s trajectory can be formally represented as \( e = (u, a_1, o_1, \dots, a_n) \), where \( u \in \mathcal{U} \) is the given task instruction, \( a \in \mathcal{A} \) is the agent action, and \( o \in \mathcal{O} \) is the observation from the environment. At each time step \( t \), the executor agent $\pi_\theta$ needs to perform implicit planning and then generate the corresponding action \( a_t \sim \pi_\theta(\cdot | u, a_1, o_1, \dots, o_{t-1})\). 
The probability of generating the task trajectory can be represented as follows:
\begin{equation}
\pi_\theta(e | u) = \prod_{t=1}^{n} \pi_\theta(a_t | u, a_1, o_1, \dots, o_{t-1}).
\end{equation}

\begin{figure*}
    \centering
    \includegraphics[width=0.94\linewidth]{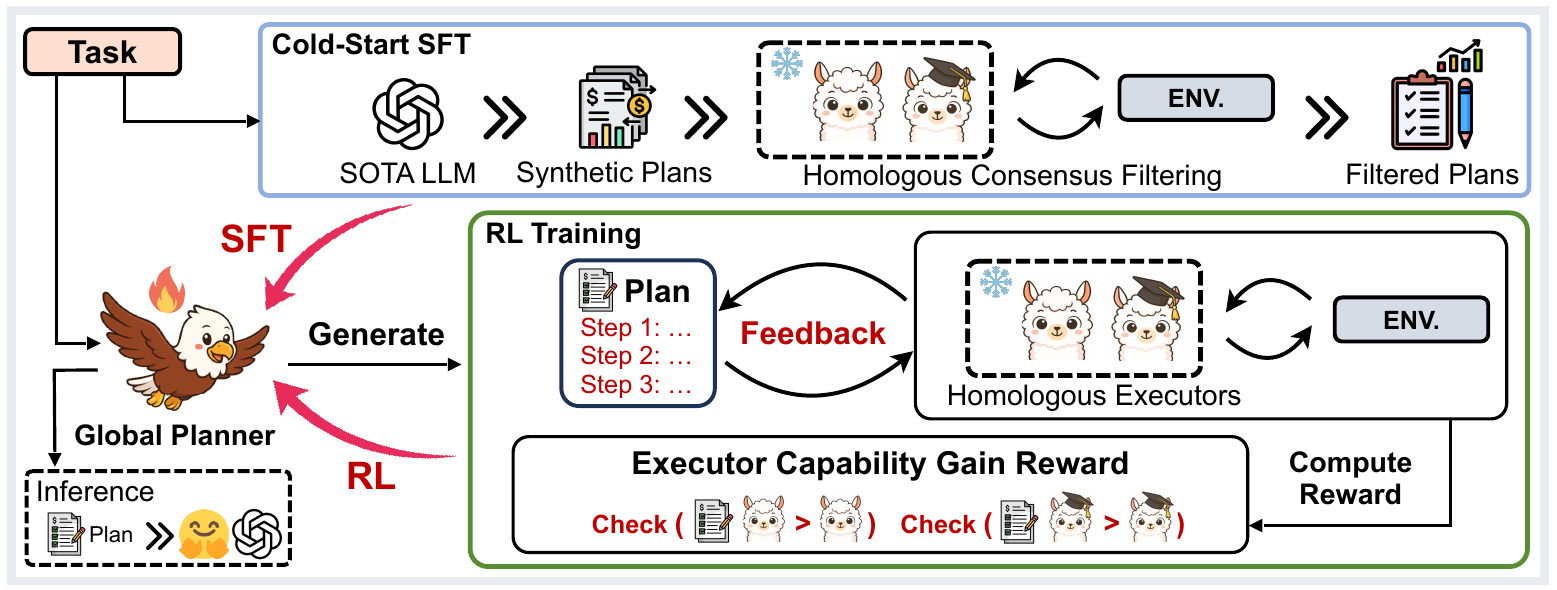}
    \caption{
     \textbf{The Overall Process of EAGLET}, including 
     (1) \textbf{Cold-Start SFT}: We synthesize high-quality global plans using the \textit{homologous consensus filtering} method for the SFT stage.
     (2) \textbf{RL Training}: We further refine the planner using a rule-based RL approach with the designed \textit{executor capability gain reward}.
    }
    \label{figure:model}
\end{figure*}

Different from implicit planning methods, which regard the agent as just an executor and focus on local planning, our framework attempts to utilize explicit knowledge to support global planning.
Inspired by \citet{qiao2024agent, xiong2025mpoboostingllmagents}, we introduce a trainable global planner $\pi_g$ to provide high-level plans.
For each task, the trained planner $\pi_g$ generates the corresponding global plan, which is then provided along with the task instruction for the executor agent to improve the global planning.
After incorporating the generated plan $p$, the probability of the executor agent $\pi_\theta$ generating the trajectory $e$ can be formulated as:
\begin{equation}
\pi_\theta(e | u, p) = \prod_{t=1}^{n} \pi_\theta(a_t | u, p, a_1, \dots, o_{t-1})\pi_g(p | u).
\end{equation}

In this way, the trained planner provides global foresight to mitigate planning hallucinations for executor agents.
The final reward \( r(u, e) \in [0, 1] \) indicating the task completion rate is calculated.


\section{Methodology}

In this paper, we propose an efficient and effective framework \textsc{EAGLET} that equips a tunable planner to explicitly endow the executor agent with global foresight, thus mitigating planning hallucinations.
For each task, the trained planner generates a global plan and gives it, along with the task instruction, to the executor agent to help improve global planning.
As shown in Figure \ref{figure:model}, the proposed \textsc{EAGLET} framework includes two key components to train an expert-level planner  without human efforts:
(1) A global plan data synthesis pipeline that generates high-quality data and the SFT stage as a cold start (\S~\ref{sft_stage});
(2) A rule-based RL stage to further enhance the global planner in handling task instructions of varying difficulty (\S~\ref{reward}).

\subsection{Cold-Start Supervised Fine-tuning}
\label{sft_stage}
To equip the global planner with the foundational capabilities to generate plans based on task instructions, we start by training the model via the SFT stage.
Previous works \citep{qiao2024agent, xiong2025mpoboostingllmagents} often rely on manually designed data or manually constructed knowledge bases to introduce explicit global guidance, making them difficult to transfer across different environments.
In this paper, we propose a global plan data synthesis pipeline that automatically generates high-quality data for the SFT stage without human effort.

\subsubsection{Data Synthesis}
Existing agent datasets only provide golden task trajectories without corresponding plans. 
Thus, we first need to construct an SFT training dataset for plan generation. 
To achieve this, we leverage the advanced reasoning LLM, e.g., GPT-5 \citep{gpt-5} and DeepSeek-V3.1-Think \citep{deepseekai2025deepseekv3technicalreport}, to synthesize the plan and thinking process.
We provide the LLM with the task instruction $u$, the trajectory $e$, and a one-shot demonstration with a crafted high-quality plan as the prompt, allowing it to provide its chain-of-thought and a plan $p$ from the trajectory, and ensuring the synthesized plans are high-level and effective.
In this way, we can obtain the thinking process and the global plan for the cold-start SFT stage.
The used prompts are shown in the Appendix \ref{appendix_prompt}.

\subsubsection{Homologous Consensus Filtering}
\label{HCF}
However, even if we apply well-designed prompts and in-context learning techniques, these synthesized plans without quality control could still be noisy or useless.
Thus, we propose the \textit{homologous consensus filtering (HCF)} strategy to avoid low-quality plan samples without human effort.
Specifically, HCF aims to filter out synthetic plans that are useless for task completion by homogeneous execution agents.
These homologous executor models \citep{yu2024language, si-etal-2025-gateau} share the same pre-training stage and model architecture, e.g., expert-level GiGPO-Llama-3.1 \citep{feng2025groupingrouppolicyoptimizationllm} and novice-level Llama-3.1 \citep{llama3}; the only difference lies in their capabilities to complete specific tasks learned from the post-training stage.
In this way, we can more comprehensively consider the effectiveness and generalization of the synthesized plans, ensuring these plans can support executor agents with different skill levels.
By using homologous executor models, we also avoid having the measurement of plan quality biased by factors other than the executor's problem-solving skills, such as context windows and parameters.

For each task instruction, we introduce two homologous executor agents, and both of them perform inference twice: once with the generated plan and once without it. 
Suppose the plan degrades the performance of either the expert-level executor agent $\hat\pi_{\theta}$ or novice-level executor agent $\hat\pi_{\tau}$. 
In that case, we consider the plan as a low-quality sample and remove it from our training set. 
Formally,
\begin{align}
\begin{split}
    F_{\textit{quality}}(p) = \mathbb I \Bigl\{ & r(u,e_{p;\hat\pi_{\theta}})\geq r(u,e_{\hat\pi_{\theta}}) \,\, \land \\
    & r(u,e_{p;\hat\pi_{\tau}})\geq r(u,e_{\hat\pi_{\tau}}) \Bigr\},
\end{split}
\end{align}
where $u$ is the task instruction, $p$ is the generated plan, $e$ is the trajectory from expert-level executor $\hat\pi_{\theta}$ or novice-level executor $\hat\pi_{\tau}$, \( r(u, e) \in [0, 1] \) is the task completion rate, and $\mathbb I$ is the indicator function for filtering low-quality plans that do not match the target.
Thus, we can filter the low-quality plan samples and the corresponding thinking processes that hinder the task execution for executor agents without relying on manual effort.

We then fine-tune the model on quality-checked training data $\mathcal D$, including both the plan $p$ and corresponding thinking process $t$ as cold start, to get the initialized global planner $\pi_g$:
\begin{equation}
    \mathcal L_\mathrm{SFT} = - \mathbb E_{(u,t,p)\sim\mathcal D}[\log\pi_g(t,p|u)].
\end{equation}

In this way, the SFT-initialized planner $\pi_g$ is equipped with capabilities to generate plans based on task instructions to provide global guidance.

\subsection{Global Planner RL Training}
\label{reward}
The SFT-initialized planner can easily memorize the simple samples in the constructed training data and struggles to generalize to harder samples.
To further enhance the global planner in handling task instructions of varying difficulty, we frame it as a rule-based RL problem and propose a well-designed reward signal called executor capability gain reward, which evaluates the gain a plan brings to executors at different capability levels.

\subsubsection{Executor Capability Gain Reward}
Having a well-designed reward is key to the effectiveness of RL training \citep{shao2024deepseekmathpushinglimitsmathematical,kimik2openagentic, si2025teaching}. 
An intuitive method is to use the plan as the input to guide the executor and then use the task completion rate as the reward to optimize the planner model.
However, this method fails to quantify the gain in task execution over the baseline without utilizing the global plan.
Meanwhile, such a reward based on a single executor agent is heavily influenced by the varying capabilities of the executor, which in turn may introduce false patterns into the optimized planner policy.
For instance, even with a low-quality plan, a capable executor agent can still use its own planning ability to accomplish the task, thereby providing a high reward for the low-quality plan.

To tackle these challenges, the proposed \textit{executor capability gain reward (ECGR)} measures the quality of the candidate plan by checking if it can help multiple homologous executor agents with different capability levels to complete tasks better within fewer interactions.
For each task instruction and generated plan, we introduce two homologous executor agents with different task execution capabilities, and both of them perform inference twice: once with the generated plan and once without it, to compute our designed reward.
By comparing the task completion rate from executions with and without the plan, we can evaluate the effectiveness and gains brought by the plan.
Meanwhile, we further apply a decay factor to encourage the planner to generate plans that motivate the executor to produce shorter and more effective trajectories, thus preventing the executor from planning hallucinations and brainless trial and error.
Formally,
\begin{align}
\small
R(p, \pi_{\theta}) =
\begin{cases}
     1 & \text{if } ~r(u,e_{p;\pi_{\theta}})> r(u,e_{\pi_{\theta}}), \\
     0 &\text{otherwise},
\end{cases} \\
\hat R(p, \pi_{\theta}) = R(p, \pi_{\theta}) \cdot (1+\alpha)^{n-m},
\label{gamma-decay factor}
\end{align}
where \( r(u, e) \in [0, 1] \) is the task completion rate, $m$ indicates the number of interactions between the executor agent $\pi_{\theta}$ and the environment in the trajectory $e$ with the generated plan $p$, $n$ indicates the number of interactions without the plan, and $\alpha$ is pre-defined hyperparameter.
In this way, we can provide positive signals to the generated plan that provide effective guidance to the executor agent. 
Moreover, while two correct trajectories may receive equal task completion rate, the shorter one earns a higher reward for the corresponding plan.



Also, to avoid the bias caused by relying on distinct capabilities of the executor agent—which may limit the generalization and applicability of generated plans—we introduce two homologous executor agents that share the same pre-training stage and model architecture \citep{yu2024language}, but with different task-specific capabilities due to the post-training phase.
Compared to using any two arbitrarily selected executor models, introducing homologous models ensures the plan can support executor agents with different skill levels and circumvents bias from the different fundamental capabilities of the executors, such as context windows.
Specifically, we utilize the expert-level executor agent $\hat\pi_{\theta}$ and novice-level executor agent $\hat\pi_{\tau}$ to thoroughly consider the generalization and influence of the generated plan from the planner $\pi_{g}$:
\begin{align}
\small
R_{\textit{ECGR}} = \hat R(p, \hat\pi_{\theta}) + \hat R(p, \hat\pi_{\tau}).
\end{align}

We further include a format reward $R_\textit{{Format}}$ that encourages adherence to a predefined output structure, e.g., using <think> and <plan> tags.
Outputs that conform to this pattern receive a reward boost, thereby enhancing output consistency. 
We use string matching to evaluate whether the generated responses adhere to the format, giving a score of 1 for a match and 0 for a mismatch.

Finally, we use the sum of these two rewards as the final composite reward $R_\textit{{Final}}$ for RL training:
\begin{align}
\label{finale_reward}
R_{\textit{Final}} = R_{\textit{ECGR}} + R_{\textit{Format}}.
\end{align}

In this way, we can thoroughly assess a plan and optimize the planner model to better handle task instructions of varying difficulty.





\subsubsection{Global Planning Policy Optimization}
\label{gppo}
For RL training of LLMs, methods based on policy optimization, such as PPO \citep{ppo} and GRPO \citep{shao2024deepseekmathpushinglimitsmathematical}, have been explored. 
Given the effectiveness of GRPO in training models and its advantages over PPO, e.g., eliminating the need to train a reward model, we utilize GRPO to optimize the planner model $\pi_{g}$.

For each task $u$, the planner model generates a group of $G$ candidate global plans, $\{p_1, \dots, p_G\}$. 
Each candidate is evaluated using a designed composite rule-based reward function shown in Eq.(\ref{finale_reward}) to evaluate the quality.
When computing ECGR, the executor agents remain frozen; gradients are backpropagated only through the planner model.
GRPO utilizes the relative performance of candidates within the group to compute an advantage $A_i$ for each output plan, guiding policy updates according to the following objective:
\begin{equation}		
\label{grpo_1}
		\resizebox{0.89\hsize}{!}{$\begin{aligned}
            \mathcal L_\mathrm{GRPO}(\pi_{g}) &= \mathbb{E}_{u, \{p_i\} \sim \pi_{g_{old}}} \left[ \frac{1}{G} \sum_{i=1}^G \mathcal{L}_i - \beta \mathbb{D}_{KL}(\pi_{g} || \pi_{g_{ref}}) \right],
		\end{aligned}$}
\end{equation}
\begin{equation}		
\label{grpo_2}
		\resizebox{0.75\hsize}{!}{$\begin{aligned}
            \mathcal{L}_i &= \min \left( w_i A_i, \text{clip}(w_i, 1 - \epsilon, 1 + \epsilon) A_i \right),
		\end{aligned}$}
\end{equation}

\noindent
where $w_i = \frac{\pi_g(p_i |u)}{\pi_{g_{old}}(p_i |u)}$, $\pi_{g_{\text{old}}}$ is the policy before the update, $\pi_{g_{\text{ref}}}$ is the reference policy (i.e., the initial model), $\epsilon$ and $\beta$ are hyperparameters controlling the update step and divergence regularization, and $A_i$ is estimated advantage within the group. 


\begin{table*}[!t]
\centering
\resizebox{\linewidth}{!}{
\begin{tabular}{c | l | c c | c c | c | c}
\toprule
{\multirow{2}{*}{\textbf{Type}}} &
{\multirow{2}{*}{\textbf{Method}}} &
\multicolumn{2}{c|}{\textbf{ScienceWorld}} &
\multicolumn{2}{c|}{\textbf{ALFWorld}} & \multicolumn{1}{c|}{\textbf{WebShop}}&
{\multirow{2}{*}{\textbf{Avg}}} \\
\cmidrule(lr){3-7}
& & Seen & Unseen & Seen & Unseen & Seen & \\
\midrule
\multirow{10}{*}{\makecell{Executor Agents \\ \textbf{w/o} Training}} 
& {\small \faToggleOff} Llama-3.1-8B-Instruct & 47.7&	42.2	&22.9	&28.4&	56.3	&39.5 \\
& {\small \faToggleOff} GPT-4.1 & 76.2 &	79.9	 & 78.6	& 72.4 &70.2 	&75.5   \\
& {\small \faToggleOff} GPT-5 & 87.6  &	88.2  &	87.9 	 &83.6 	 &75.3  &	84.5    \\
& {\small \faToggleOff} DeepSeek-V3.1-Non-Think &57.4&	58.1&	50.0	&37.3&	58.6 &	52.3  \\
& {\small \faToggleOff} DeepSeek-V3.1-Think &78.7	&76.2	&81.4	&69.4	& 70.8 &	75.3  \\
& {\small \faToggleOn} MPO + Llama-3.1-8B-Instruct &56.5&	55.5&	50.0	&52.2	&63.2	&55.5  \\
& {\small \faToggleOn} MPO + GPT-4.1  & 80.4 &	83.8 &	81.4 	&79.1 &	72.5 &	79.4  \\
& {\small \faToggleOn} MPO + GPT-5 &87.8 &	89.0 &	88.2 &	83.6 &	76.6& 	85.1  \\
\cmidrule(lr){2-8}
& {\small \faToggleOn} \textbf{\ours} + Llama-3.1-8B-Instruct& 59.3 {\small \textcolor{red}{+11.6}} & 61.6 {\small \textcolor{red}{+19.4}} & 54.3 {\small \textcolor{red}{+31.4}} & 55.2 {\small \textcolor{red}{+26.8}} & 66.7 {\small \textcolor{red}{+10.4}} & 59.4 {\small \textcolor{red}{+19.9}} \\
& {\small \faToggleOn} \textbf{\ours} + GPT-4.1 &82.6 {\small \textcolor{red}{~~+6.4}} & 85.6 {\small \textcolor{red}{~~+5.7}} & 84.3 {\small \textcolor{red}{~~+5.7}} & 83.6 {\small \textcolor{red}{+11.2}} & 74.7 {\small \textcolor{red}{~~+4.5}} & 82.2 {\small \textcolor{red}{~~+6.7}}  \\
\rowcolor{blue!5} \cellcolor{white} & {\small \faToggleOn} \textbf{\ours} + GPT-5 &\textbf{89.5} {\small \textcolor{red}{~~+1.9}} 
&\textbf{90.1} {\small \textcolor{red}{~~+1.9}} & 
\textbf{90.2} {\small \textcolor{red}{~~+2.3}} & \textbf{90.7} {\small \textcolor{red}{~~+7.1}} & \textbf{80.1} {\small \textcolor{red}{~~+4.8}} & 
\textbf{88.1} {\small \textcolor{red}{~~+3.6}} \\

\midrule
\multirow{11}{*}{\makecell{Executor Agents \\ \textbf{w/} Training}}
& {\small \faToggleOff} AgentTuning &65.3&	57.0	&79.3	&71.6	& 63.3	& 67.3 \\
& {\small \faToggleOff} ETO & 81.3&	74.1&	77.1	&76.4&68.4&	75.5  \\
& {\small \faToggleOff} GiGPO &83.3	&74.5&	85.2	&88.6	&82.5	& 82.8 \\
& {\small \faToggleOn} WKM& 82.1&	76.5&	77.5	&78.2	&66.9&	76.2   \\
& {\small \faToggleOn} KnowAgent &81.7&	69.6&	80.0&	74.9&	64.8	& 74.2   \\
& {\small \faToggleOn} MPO + AgentTuning &	 70.2&	 65.9 &	80.7&	 81.3 &	65.5 &		72.7 \\
& {\small \faToggleOn} MPO + ETO &	83.4 &	80.8 &	85.0 &	79.1 &	70.2&	79.7 \\
& {\small \faToggleOn} MPO + GiGPO & 84.6 	&78.2& 	86.6 	&88.1 	&83.5 	&84.2  \\
\cmidrule(lr){2-8}
& {\small \faToggleOn} \textbf{\ours} + AgentTuning  
& 74.3 {\small \textcolor{red}{~~+9.0}} 
& 68.4 {\small \textcolor{red}{+11.4}} 
& 82.3 {\small \textcolor{red}{~~+3.0}} 
& 83.4 {\small \textcolor{red}{+11.8}} 
& 66.7 {\small \textcolor{red}{~~+3.4}} 
& 75.0 {\small \textcolor{red}{~~+7.7}} \\
& {\small \faToggleOn} \textbf{\ours} + ETO &
84.7 {\small \textcolor{red}{~~+3.4}} 
& 82.5 {\small \textcolor{red}{~~+8.4}} 
& 87.3 {\small \textcolor{red}{+10.2}}
& 83.2 {\small \textcolor{red}{~~+6.8}} 
& 72.6 {\small \textcolor{red}{~~+4.2}} 
& 82.0 {\small \textcolor{red}{~~+6.6}}  \\
\rowcolor{blue!5} \cellcolor{white} & {\small \faToggleOn} \textbf{\ours} + GiGPO & \textbf{87.7} {\small \textcolor{red}{~~+4.4}} & \textbf{83.6} {\small \textcolor{red}{~~+9.1}} & \textbf{88.6} {\small \textcolor{red}{~~+3.4}} & \textbf{91.8} {\small \textcolor{red}{~~+4.2}} & \textbf{86.2} {\small \textcolor{red}{~~+3.7}} & \textbf{87.6} {\small \textcolor{red}{~~+5.0}} \\
\bottomrule
\end{tabular}
}
\caption{
\textbf{Effectiveness Results.}
The best results are marked in \textbf{bold}.
All the explicit planning methods ({\small \faToggleOn}) use explicit knowledge to guide executors.
{\small \faToggleOff} represents prompt engineering methods and implicit planning methods without global foresight.
\textcolor{red}{Red} shows the changes of \ours~relative to the the baselines without global plans.
Executor agents w/ training baselines are implemented on Llama-3.1-8B-Instruct following \citet{xiong2025mpoboostingllmagents}.
}
\label{tab:main_results}
\end{table*}

\section{Experiments}
In this section, we conduct experiments and analyses to show the effectiveness of EAGLET.

\subsection{Experiment Settings}
\textbf{Benchmarks.}
We conducted experiments on ScienceWorld \citep{wang-etal-2022-scienceworld}, ALFWorld \citep{shridhar2021alfworld}, and WebShop \citep{yao2022webshop}.
ScienceWorld presents a challenging benchmark for complete scientific experiments in a highly interactive environment.
ALFWorld contains household tasks that require agents to explore rooms and accomplish objectives.
WebShop is a web-based interactive environment designed to test agents in realistic online shopping scenarios.
Both ALFWorld and ScienceWorld include seen and unseen scenarios to assess in-distribution and out-of-distribution generalization of the agents separately.
In contrast, WebShop only contains seen scenarios for testing.
More details are shown in Appendix \ref{appendix_dataset}.

\noindent
\textbf{Baselines.}
We fully compare the proposed method with a series of  baselines, including: 
(1) \textbf{Closed-source LLMs}: We test four state-of-the-art models, namely GPT-4.1 \citep{OpenAI2023GPT4TR}, GPT-5 \citep{gpt-5}, DeepSeek-V3.1-Non-Think \citep{deepseekai2025deepseekv3technicalreport}, and DeepSeek-V3.1-Think, which represent advanced capabilities in reasoning and understanding.
(2) \textbf{Implicit planning methods}: These works enhance agent planning capabilities via parameter updates. 
AgentTuning \citep{zeng-etal-2024-agenttuning} uses SFT on collected trajectories to improve the task-solving capabilities.
ETO \citep{song-etal-2024-trial} further teaches executor agents to learn from failed trajectories using an exploration-based trajectory optimization method.
GiGPO \citep{feng2025groupingrouppolicyoptimizationllm} proposes a novel outcome-based RL algorithm that achieves fine-grained reward assignment for training the executor agent.
(3) \textbf{Explicit planning methods}: These methods try to use explicit knowledge to guide executor agents.
WKM \citep{qiao2024agent} and KnowAgent \citep{zhu-etal-2025-knowagent} collect the prior knowledge and incorporate external action knowledge into the executor agent training.
MPO \citep{xiong2025mpoboostingllmagents} trains a planner model to provide the global guidance via applying SFT on human-modified data and DPO training.

\noindent
\textbf{Metrics and Evaluation.}
ScienceWorld and WebShop provide dense final rewards ranging from 0 to 1 to measure the completion level of the task, whereas ALFWorld offers only binary rewards indicating whether the agent has completed the task.
For all the datasets, we apply \textbf{average reward} as the metric, which calculates the mean reward across all task instances. 
We also report the success rate in Appendix \ref{appendix_eval_imp}. 
For evaluation, we apply ReAct \citep{yao2023react} prompting with a one-shot in-context example following \citet{xiong2025mpoboostingllmagents} to leverage these models’ abilities.
However, GiGPO is an exception: we use the prompt from the original paper \citep{feng2025groupingrouppolicyoptimizationllm} as the GiGPO method does not employ ReAct-style prompts during RL training.

\begin{table}[!t]
\centering
\resizebox{0.98\linewidth}{!}{
\begin{tabular}{lccccc}
\toprule
\makecell[l]{\textbf{Method}} 
& \makecell{\textbf{Plug-and-Play}\\\textbf{ Manner}}
& \makecell{\textbf{Explicit}\\\textbf{Guidance}}
& \makecell{\textbf{Human-effort}\\\textbf{Free}} &\makecell{\textbf{Data}\\\textbf{Efficient}}
& \makecell{\textbf{RL Training}\\\textbf{\# Iterations}} \\
\midrule
AgentTuning & \textcolor{red!90!black}{$\times$} 
 & \textcolor{red!90!black}{$\times$} & \textcolor{green!80!black}{\checkmark}  & \textcolor{red!90!black}{$\times$} & N/A \\
ETO & \textcolor{red!90!black}{$\times$} & \textcolor{red!90!black}{$\times$}  & \textcolor{green!80!black}{\checkmark}  & \textcolor{green!80!black}{\checkmark} & N/A \\
KnowAgent & \textcolor{red!90!black}{$\times$} & \textcolor{green!80!black}{\checkmark} & \textcolor{red!90!black}{$\times$}& \textcolor{green!80!black}{\checkmark} & N/A \\
WKM & \textcolor{red!90!black}{$\times$} & \textcolor{green!80!black}{\checkmark} & \textcolor{green!80!black}{\checkmark} & \textcolor{green!80!black}{\checkmark} & N/A \\
MPO & \textcolor{green!80!black}{\checkmark} & \textcolor{green!80!black}{\checkmark} & \textcolor{red!90!black}{$\times$} & \textcolor{red!90!black}{$\times$} & N/A \\
GiGPO  & \textcolor{red!90!black}{$\times$} & \textcolor{red!90!black}{$\times$}  & \textcolor{green!80!black}{\checkmark} & \textcolor{green!80!black}{\checkmark} & \textasciitilde 400 (Inefficient) \\
\rowcolor{blue!5} \textbf{\ours} & \textcolor{green!80!black}{\checkmark} & \textcolor{green!80!black}{\checkmark} & \textcolor{green!80!black}{\checkmark}& \textcolor{green!80!black}{\checkmark} & \textbf{\textasciitilde 50 (Efficient)} \\
\bottomrule
\end{tabular}}
\caption{
\textbf{Efficiency Results.} 
``Data Efficient'' shows that this method does not require data beyond the original training set.
``N/A'' indicates that the method does not involve online RL training.}
\label{tb:efficiency} 
\end{table}
\begin{table}[!h]
\centering
\scriptsize
\resizebox{0.98\linewidth}{!}{
\begin{tabular}{l c c c c c}
\toprule
{\multirow{2}{*}{\textbf{Method}}} &
\multicolumn{2}{c}{\textbf{ScienceWorld}} &
\multicolumn{2}{c}{\textbf{ALFWorld}} &
{\multirow{2}{*}{\textbf{Avg}}} \\
\cmidrule(lr){2-5}
 & Seen & Unseen & Seen & Unseen &  \\
\midrule
\multicolumn{4}{l}{\textit{Executor Agent: GPT-4.1}} \\
- w/o Guidance      & 14.4   & 16.7   & 10.8 & 9.9 & 13.0 \\
MPO      & 13.6   & 16.5   & 10.6 & 10.6 & 12.8 \\
\rowcolor{blue!5} \textbf{\ours} & \textbf{12.2} & \textbf{14.3} & \textbf{9.4} & \textbf{8.6} & \textbf{11.1} \\
\midrule
\multicolumn{4}{l}{\textit{Executor Agent: GPT-5}} \\
- w/o Guidance &  11.3 & 13.1 & 10.4 & 10.7 & 11.4 \\
MPO &  12.1 & 15.5 & 9.7 & 9.9 & 11.8  \\
\rowcolor{blue!5} \textbf{\ours} & \textbf{10.2} & \textbf{10.6} & \textbf{8.6} & \textbf{8.2} & \textbf{9.4} \\
\bottomrule
\end{tabular}
}
\caption{
\textbf{Average Steps.}
We report the average steps from executors to finish tasks under different guidance.
}
\label{tab:step}
\end{table}

\noindent
\textbf{Implementation Details.}
For a fair comparison, we use Llama-3.1-8B-Instruct \citep{llama3} to train the planner following \citet{xiong2025mpoboostingllmagents}.
For the homologous consensus filtering and computing executor capability gain reward, we select Llama-3.1-8B-Instruct and GiGPO-Llama-3.1-8B as homologous executors.
We utilize DeepSeek-V3.1-Think instead of GPT-5 to synthesize plans and the thinking process for the SFT stage, as we can not access the thinking process from GPT-5.
More details are shown in Appendix \ref{appendix_eval_imp}.

\subsection{Results}
\textbf{Effectiveness Results.}
As shown in Table \ref{tab:main_results}, the incorporation of \ours-generated plans consistently improves executor agent performance.
Executor agents equipped with our planner greatly outperform those without it, achieving new state-of-the-art results.
Different from utilizing explicit knowledge during the executor training stage, like WKM and KnowAgent, our plug-and-play planner can benefit both trained models and closed-sourced LLMs, showing its flexibility.
For the unseen parts of ScienceWorld and ALFWorld, our planner can also generalize to them and generate high-quality plans, demonstrating strong generalizability.
Meanwhile, compared with MPO, due to the well-designed RL stage to enhance the planner in handling task instructions of varying difficulty, our method can better help skilled executors to complete tasks, e.g., GPT-5 and GiGPO.


\begin{table}[t]
\centering
\scriptsize
\begin{tabular}{l c c c}
\toprule
{\multirow{2}{*}{\textbf{Method}}} &
\multicolumn{2}{c}{\textbf{ALFWorld}} &
{\multirow{2}{*}{\textbf{Avg}}} \\
\cmidrule(lr){2-3}
 & Seen & Unseen &  \\
\midrule
Using GPT-4.1 as Planner & 80.4 & 78.5 & 79.5 \\
\hdashline[2pt/3pt]
\rowcolor{blue!5} \textbf{\ours} & \textbf{84.3} & \textbf{83.6} &  \textbf{84.0} \\
- w/o Guidance from \ours & 78.6 	& 72.4  & 75.5 \\
- w/o Cold-Start SFT     &   79.3 &  74.6  & 77.0 \\
- w/o Homologous Consensus Filtering   & 82.1   & 81.2   &  81.7 \\
- w/o Global Planner RL Training & 80.7 & 78.5 & 79.6 \\
- w/o Executor Capability Gain Reward    &  82.3  & 80.6    & 81.5 \\

\bottomrule
\end{tabular}
\caption{
\textbf{Ablation Study.}
We use GPT-4.1 as the executor agent to report the results.
}
\label{tab:ablation}
\end{table}
\begin{table}[t]
\centering
\scriptsize
\begin{tabular}{l l c c c}
\toprule
{\multirow{2}{*}{\textbf{Executor}}} & 
{\multirow{2}{*}{\textbf{Type}}} &
\multicolumn{2}{c}{\textbf{ALFWorld}} &
{\multirow{2}{*}{\textbf{Avg}}} \\
\cmidrule(lr){3-4}
 & & Seen & Unseen &  \\
\midrule
\multirow{3}{*}{GPT-4.1} 
& \cellcolor{blue!5}Instruction & \cellcolor{blue!5}\textbf{84.3} & \cellcolor{blue!5}\textbf{83.6} & \cellcolor{blue!5}\textbf{84.0} \\
 & Thought& 83.6   & 81.7   & 82.7   \\
 & Observation  & 82.7  & 82.1   & 82.4   \\
\midrule
\multirow{3}{*}{GPT-5} 
 & \cellcolor{blue!5} Instruction & \cellcolor{blue!5}\textbf{90.2}  & \cellcolor{blue!5}\textbf{90.7} & \cellcolor{blue!5}\textbf{90.5} \\
 & Thought & 89.5   & 89.6   & 89.6   \\
 & Observation  & 89.6  & 87.3   & 88.5   \\
\bottomrule
\end{tabular}
\caption{
\textbf{Exploration of Positions.}
The impact of different plan insertion positions on agent performance.
}
\label{tab:pos}
\end{table}

\noindent
\textbf{Efficiency Results.}
Our \ours~achieves better efficiency in both training and task execution.
As shown in Table \ref{tb:efficiency}, we introduce an efficient planner to provide explicit guidance without human effort.
Compared with the explicit planning methods, our \ours~do not rely on any human effort.
Meanwhile, our method introduces a plug-and-play planner, avoiding the need to retrain the executor agent.
Compared with data-inefficient SFT-based methods such as AgentTuning and WKM, our approach does not require introducing additional annotated data beyond the original training set to generalize well.
Compared to RL-based methods like GiGPO, which are hampered by the difficult credit assignment problem for individual steps arising from sparse and delayed rewards, our method achieves superior performance with greater training efficiency.
Also, under the generated plans from \ours, executor agents can take fewer steps to finish more tasks, as shown in Table \ref{tab:step}, achieving better efficiency and performance in task execution.

\noindent
\textbf{Ablation Study.}
We also conduct the ablation study as shown in Table \ref{tab:ablation}.
Specifically, we systematically remove or modify key components of our \ours~framework to understand their individual contributions to overall performance.
When we individually remove the plans generated by \ours, the cold-start SFT stage, or the RL training stage, we observe a significant drop in performance.
`` -w/o Homologous Consensus Filtering'' represents that we use unfiltered plans for the SFT stage.
`` -w/o Executor Capability Gain Reward'' means we use the plan to guide the executor Llama-3.1-8B-Instruct and then use the task completion rate as the reward to optimize the planner.
The results show the effectiveness of our designed components.
More fine-grained variant methods testing can be found in Appendix \ref{appendix_var_test}, e.g., the rationale for using homologous executors instead of any two arbitrarily selected models in our proposed approaches.
\subsection{Analysis}

\begin{table}[t]
\centering
\scriptsize
\begin{tabular}{l l c c c}
\toprule
{\multirow{2}{*}{\textbf{Executor}}} & 
{\multirow{2}{*}{\textbf{Type}}} &
\multicolumn{2}{c}{\textbf{ALFWorld}} &
{\multirow{2}{*}{\textbf{Avg}}} \\
\cmidrule(lr){3-4}
 & & Seen & Unseen &  \\
\midrule
\multirow{2}{*}{Llama-3.1-8B-Instruct}
 & - w/o Guidance &   22.9 &  28.4  &   25.7 \\
& \cellcolor{blue!5}\textbf{\ours} & \cellcolor{blue!5}\textbf{54.3} & \cellcolor{blue!5}\textbf{55.2} & \cellcolor{blue!5}\textbf{55.8} \\
\midrule
\multirow{2}{*}{\makecell[c]{Llama-3.1-8B-Instruct \\+ Reflexion}}
 & - w/o Guidance &   26.5 &  33.1  &  29.8  \\
& \cellcolor{blue!5}\textbf{\ours} & \cellcolor{blue!5}\textbf{55.4} & \cellcolor{blue!5}\textbf{55.6} & \cellcolor{blue!5}\textbf{55.5} \\
\midrule
\multirow{2}{*}{Llama-3.1-70B-Instruct}
 & - w/o Guidance &  78.6  &  73.9  &  76.3  \\
& \cellcolor{blue!5}\textbf{\ours} & \cellcolor{blue!5}\textbf{87.9} & \cellcolor{blue!5}\textbf{88.2} & \cellcolor{blue!5}\textbf{88.1} \\
\midrule
\multirow{2}{*}{Qwen2.5-7B-Instruct}
 & - w/o Guidance &  71.4  &  75.4  & 73.4   \\
& \cellcolor{blue!5}\textbf{\ours} & \cellcolor{blue!5}\textbf{83.2} & \cellcolor{blue!5}\textbf{84.5} & \cellcolor{blue!5}\textbf{83.9} \\
\bottomrule
\end{tabular}
\caption{
\textbf{Generalization Across Executors.}
The impact of different executor backbones and prompting methods.
}
\label{tab:diff_executor}
\end{table}

\begin{figure}[t]
    \hspace{-5mm}
    \centering
    \includegraphics[width=0.3\textheight]
    {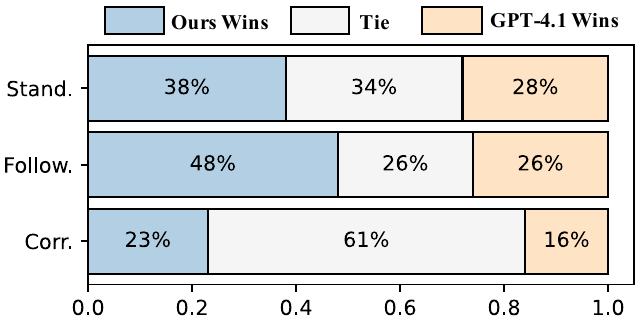}
    \caption{
    \textbf{Plan Quality Analysis}. 
    The comparison of GPT-4.1-generated and our plans on ALFWorld.}
    \label{fig_gpt}
\end{figure}

\noindent
\textbf{The Impact of Plan Insertion Position.}
We further investigate the impact of different insertion positions on performance: in the task instruction (used in our main experiments), in the executor’s thought process, and in the environment observation.
Shown in Table \ref{tab:pos}, we find that although different insertion positions can all lead to performance improvements, insertion into the task instruction consistently yields the best performance.

\noindent
\textbf{Generalization Across Executor Agents.}
We explore the impact of different executor backbones and prompting strategies shown in Table \ref{tab:diff_executor}.
We can observe that our method is effective not only across different executor backbones (Llama-3.1 vs. Qwen-2.5 \citep{qwen2.5}), but also brings performance improvements to models with different parameter sizes (Llama-3.1-8B vs. Llama-3.1-70B).
Similarly, our method is compatible with different prompting strategies. 
We can find that when the executor model uses different prompting strategies like Reflexion \citep{shinn2023reflexion}, our method can also lead to significant performance improvements.

\begin{table}[t]
\centering
\scriptsize
\begin{tabular}{l c c c}
\toprule
{\multirow{2}{*}{\textbf{Foundational Models}}}  &
\multicolumn{2}{c}{\textbf{ALFWorld}} &
{\multirow{2}{*}{\textbf{Avg}}} \\
\cmidrule(lr){2-3} & Seen & Unseen &  \\
\midrule
\multicolumn{3}{l}{\textit{Executor Agent: GPT-4.1}} \\
-w/o Guidance & 78.6 & 72.4  & 75.5 \\
Using GPT-4.1 as Planner & 80.4 & 78.5 & 79.5 \\
\hdashline[2pt/3pt]
EAGLET + Llama-3.1-8B-Instruct & 84.3 & 83.6 & 84.0 \\
EAGLET + Qwen-2.5-7B-Instruct & 83.8 & 81.5 & 82.7 \\
EAGLET + Qwen-2.5-14B-Instruct & \textbf{84.6} & \textbf{84.1} & \textbf{84.4} \\
\bottomrule
\end{tabular}
\caption{
\textbf{Generalization Across Foundational Models.}
The impact of different backbones of the trained planner.
}
\label{tab:found_models}
\end{table}

\begin{table}[!t]
\centering
\scriptsize
\resizebox{0.98\linewidth}{!}{
\begin{tabular}{l c c c c}
\toprule
{\multirow{2}{*}{\textbf{Method}}} &
\multicolumn{2}{c}{\textbf{Reward}} &
\multicolumn{2}{c}{\textbf{Steps}} \\
\cmidrule(lr){2-5}
 & Seen & Unseen & Seen & Unseen \\
\midrule
- w/o Guidance & 78.6 & 72.4 &  10.8 & 9.9  \\
MPO & 81.4 & 79.1 & 10.6 & 10.6 \\
- w/o Gamma-decay Factor ($\alpha$ = 0) & 83.4 & 83.0 & 10.4 & 9.5 \\
\rowcolor{blue!5} \textbf{\ours} ($\alpha$ = 0.2) & 84.3 & 83.6 & 9.4 & 8.6 \\
- w/ $\alpha$ = 0.05  & 83.6 & 83.2 & 10.0 &9.0  \\
- w/ $\alpha$ = 0.5 & 83.2 & 83.0 & 9.3 & 8.9 \\
- w/ $\alpha$ = 1 & 82.9 & 82.8 & 9.2 & 8.5 \\
\bottomrule
\end{tabular}
}
\caption{
\textbf{Parameter Study.}
We report the average reward and the average step
from the executor agent GPT-4.1 to finish the task on ALFWorld.
}
\label{tab:para}
\end{table}

\noindent
\textbf{Plan Quality Analysis.}
We delve into whether the generated plans based on \ours~align with known characteristics of high-quality plans shown in Figure \ref{fig_gpt}.
We conduct the pair-wise evaluation between our and GPT-4.1-generated plans from three perspectives: correctness, followability, and standardization.
We test the plans from both seen and unseen scenarios from ALFWorld using GPT-4.1 as a judge, and the used prompt can be found in Appendix \ref{appendix_prompt}.
We can find that our plans outperform GPT-4.1-generated ones across all three dimensions. 
The advantages in followability make it easier for the agent to effectively plan and execute tasks, leading to higher task completion rates.

\noindent
\textbf{Generalization Across Foundational Models.}
As shown in Table \ref{tab:found_models}, we can find that using different foundational models to train our planner can consistently improve the performance.
Meanwhile, our trained planner outperforms the GPT-4.1-based planner, demonstrating the necessity of training planners specifically for particular tasks.

\noindent
\textbf{Parameter Study}
\label{appendix_para_study}
We introduce a gamma-decay factor $\alpha$ in Eq.(\ref{gamma-decay factor}) to encourage the planner to generate plans that motivate the executor to produce shorter and more effective trajectories.
This is the only extra hyperparameter in our method, and we further conduct a parameter study to explore the robustness of our method.
We observe from Table~\ref{tab:para} that the introduction of the gamma-decay factor $\alpha$ substantially enhances both task success and efficiency. 
A moderate setting ($\alpha=0.2$) yields the best trade-off, achieving the highest average reward while notably reducing execution steps on both seen and unseen tasks. 
Smaller values (e.g., $\alpha=0.05$) bring marginal improvements, whereas larger values (e.g., $\alpha \geq 0.5$) overly emphasize trajectory brevity at the expense of plan quality. 
Meanwhile, as long as $\alpha$ is properly set, it will always lead to performance improvement, which demonstrates the effectiveness of our design.
These results demonstrate the robustness of our method and highlight the importance of $\alpha$ in balancing correctness and efficiency during global planning.


\section{Related Work}
With the recent progress in LLMs \citep{Wang_2024, wei2022emergent}, researchers attempt to use prompting techniques to build agents that can handle real-world tasks \citep{wei2022chain, song2023restgptconnectinglargelanguage,koh2024treesearchlanguagemodel, lei2025rhinoinsight, zhao-etal-2025-looking, si2025faithlens}, e.g., writing code \citep{qian-etal-2024-chatdev} and web navigation \citep{chae2025web}.
However, these agents with static workflow struggle with planning hallucinations, leading to brainless trial-and-error in the environment \citep{zhu-etal-2025-knowagent}.
Thus, some recent works attempt to improve the planning capabilities of LLM-based agents by applying the SFT stage on collected expert trajectories \citep{zeng-etal-2024-agenttuning, song-etal-2024-agentbank}, while others enable agents to interact with the environment and leverage reinforcement learning to learn from failed explorations \citep{song-etal-2024-trial,fei2025unleashingembodiedtaskplanning, hu2025divide, feng2025groupingrouppolicyoptimizationllm}. 
However, these \textit{implicit planning methods} require agents to learn from collected data or agent-environment interactions implicitly, and focus solely on local planning, i.e, the planning that occurs through interleaved reasoning and action generation, limiting the global planning ability for long-horizon tasks.
Also, implicit planning methods are generally inefficient: SFT-based methods are data-inefficient, requiring substantial amounts of expert-annotated data to generalize well due to limited exploration of environments; RL-based methods are training-inefficient, requiring extensive training time and iterations to converge, as agent-environment interactions involve many decision steps, sparse rewards, and long-term credit assignment.
Such implicit methods often require retraining each time a new agent is deployed, and need a large amount of expert-annotated trajectories or more training iterations to converge, making them inefficient.
To address the challenges of myopic reasoning and planning hallucinations \citep{zhu-etal-2025-knowagent}, some works use language models to synthesize task-related knowledge \citep{zhou2023agentsopensourceframeworkautonomous, ye2023proagentroboticprocessautomation, fu2024autoguideautomatedgenerationselection}, but the generated knowledge is static and cannot be further optimized through agent-environment feedback, leading to suboptimal performance. 
Recently, some approaches \citep{guan2024amor, li2024formalllmintegratingformallanguage, 10.1609/aaai.v38i17.29936, qiao2024agent,zhu-etal-2025-knowagent,xiong2025mpoboostingllmagents} have explored using explicit knowledge to guide task execution. 
However, these \textit{explicit planning methods} often require manually designed training data or manually constructed knowledge bases, making them difficult to transfer across different environments. 
In contrast, our proposed \textsc{Eaglet} presents a novel framework that is both efficient and effective by training a global planner through a fully automated pipeline, thus removing the dependency on manual efforts.
This results in a plug-and-play planner that provides global foresight to mitigate planning hallucinations for any new executor agent, therefore achieving state-of-the-art performance.
\section{Conclusion}
\label{subsection:conclusion}
In this paper, we introduce EAGLET, an efficient and effective method to train a global planner for long-horizon agent tasks. 
Specifically, we first synthesize high-quality plans from an advanced LLM using our proposed homologous consensus filtering and apply cold-start SFT. 
Then, we further enhance the global planner with a rule-based RL stage using our executor capability gain reward, ensuring it can handle task instructions of varying difficulty.
The resulting planner is plug-and-play and mitigates planning hallucinations, leading to state-of-the-art performance on three long-horizon agent benchmarks. 
EAGLET is also efficient in both the training stage and task execution stage, and does not need any manual effort. 
These findings underscore the significant potential of our approach to enhance LLM-based agent planning capabilities.

\section*{Acknowledgements}
We would like to thank the anonymous reviewers for their suggestions.
This work is supported by the National Natural Science Foundation of China (No.T2341003), National Natural Science Foundation of China (No. 62236011), and a grant from the Guoqiang Institute, Tsinghua University.

\section*{Limitations}
While our proposed EAGLET framework demonstrates strong effectiveness and efficiency across multiple long-horizon agent benchmarks, several directions remain unexplored. 
First, our experiments primarily focus on text-based interactive environments. 
Extending EAGLET to multi-modal settings may present additional challenges. 
Second, although our homologous consensus filtering and executor capability gain reward provide a principled way to train without human effort, they still rely on the availability of diverse executor agents. 
Investigating more lightweight or self-improving evaluation strategies could broaden applicability. 
Also, we utilize the open-source executor agents with different task-specific capabilities (i.e., Llama-3.1-8B-Instruct and  GiGPO-Llama-3.1-8B in our main experiments) instead of API-based LLMs, e.g., GPT-4.1 and GPT-5.
This is because API-based LLMs introduce extra training overhead, including both training time and expensive API costs.
We regard such experiments as future work.
Finally, while our planner generalizes well across unseen scenarios and executor backbones, studying long-term transfer across domains and tasks of significantly different structures remains an open question. 
We view these aspects as promising avenues for future research.



\bibliography{custom}

\appendix
\section*{Appendix}

\noindent This appendix is organized as follows.  

\begin{itemize}
    \item In Section~\ref{appendix_dataset}, we show the details of the used datasets, including ScienceWorld, ALFWorld, and WebShop.
    \item In Section~\ref{appendix_eval_imp}, we describe the details of implementation and evaluation.
    \item In Section~\ref{appendix_var_test}, we conduct fine-grained variant method testing to validate the effectiveness of our proposed designs.
    \item In Section~\ref{appendix_case_study}, we provide a case study from ALFWorld to illustrate the advantages of our approach compared with baselines.
    \item In Section~\ref{appendix_prompt}, we list the prompts used in our data synthesis, training, and evaluation.
\end{itemize}


\section{Dataset Details}
\label{appendix_dataset}
\textbf{ScienceWorld.}
ScienceWorld \citep{wang-etal-2022-scienceworld} is a text-based virtual environment designed as a rigorous evaluation platform, with a specific focus on assessing and advancing scientific reasoning capabilities. 
This environment enables researchers to systematically measure the performance of agents within open-ended, complex scenarios. 
ScienceWorld includes tasks derived from standard elementary science curricula, spanning topics such as phase transitions, measurement, electricity, life sciences, plant development, chemical reactions, and genetics. Agents are placed in an embodied, interactive setting, requiring them to comprehend and operationalize sophisticated scientific concepts. 
Each task in ScienceWorld is composed of multiple subgoals, and final rewards are determined based on the achievement of these subgoals. 
ScienceWorld test set features previously unseen task variants—for instance, while training tasks might require boiling water, test tasks may involve boiling lead. 
In line with prior work \citep{song-etal-2024-trial, xiong2025mpoboostingllmagents}, we utilize the original test set with 241 samples to assess the generalization capacity of our planner in unseen scenarios, and employ the original validation set with 194 samples to evaluate performance in seen settings.

\noindent
\textbf{ALFWorld.}
ALFWorld encompasses six categories of planning tasks set primarily in home environments, covering not only basic object manipulation (such as pick and place) but also tasks requiring complex interaction sequences. 
For example, the heating task requires models to first identify target objects, move them to heating devices (like a microwave), execute the heating operation, and finally place them in designated locations to complete the task.
Following prior research \citep{song-etal-2024-trial, xiong2025mpoboostingllmagents}, we evaluate model performance under two conditions: seen and unseen scenarios. 
Seen scenarios consist of 140 task instances from rooms encountered during training; unseen scenarios comprise 134 task instances from entirely new rooms with container arrangements and scene organizations distinctly different from training tasks, designed to evaluate the model’s zero-shot generalization capabilities.

\noindent
\textbf{WebShop.}
WebShop is a complex, web-based interactive environment designed to test the LLM agents in realistic online shopping scenarios. 
To complete the task, the agent must interact with a simulated HTML-based shopping website to search for, navigate to, and ultimately purchase a suitable item. 
Once the agent clicks the "buy" option, the environment provides a final reward, which is calculated based on the matching heuristics of the product’s attributes and price.
It contains over 1.1 million products, providing a rich and diverse action space.
Following \citet{song-etal-2024-trial, xiong2025mpoboostingllmagents}, we use the same 200 samples to conduct the experiments to make a fair comparison.


\section{Implementation and Evaluation Details}
\label{appendix_eval_imp}

\begin{table}[!t]
\centering
\scriptsize
\resizebox{0.98\linewidth}{!}{
\begin{tabular}{l c c c c}
\toprule
{\multirow{2}{*}{\textbf{Method}}} &
\multicolumn{2}{c}{\textbf{ScienceWorld}} &
\multicolumn{1}{c}{\textbf{WebShop}} &
\multicolumn{1}{c}{\textbf{Avg}}  \\
\cmidrule(lr){2-4}
 & Seen & Unseen & Seen  \\
\midrule
\multicolumn{4}{l}{\textit{Executor Agent: GPT-4.1}} \\
- w/o Guidance  &  74.2	& 74.3 & 61.0 & 69.8 \\
MPO    & 77.8	& 82.4  & 63.5  & 74.6 \\
\rowcolor{blue!5} \textbf{\ours} & \textbf{79.3} & \textbf{83.8} & \textbf{66.0} & \textbf{76.4} \\
\midrule
\multicolumn{4}{l}{\textit{Executor Agent: GPT-5}} \\
- w/o Guidance & 75.8 & 	80.1 & 64.0 & 73.3 \\
MPO & 75.8	& 80.5 & 65.0 & 73.8 \\
\rowcolor{blue!5} \textbf{\ours} & \textbf{80.4} &	\textbf{82.1} & \textbf{70.0} & \textbf{77.5} \\
\bottomrule
\end{tabular}
}
\caption{
\textbf{Success Rates.}
We report the success rates on ScienceWorld and WebShop benchmarks.
}
\label{tab:success}
\end{table}
\noindent
\textbf{Hyperparameters and Devices.}
For SFT training, we use the Adam optimizer \citep{kingma2017adammethodstochasticoptimization} to train our model, with a $1 \times 10^{-5}$ learning rate with a weight decay of 0.1, and a batch size of 16, steering the training across 3 epochs.
We conduct our SFT stage with DeepSpeed+ZeRO3 and BF16.
For RL training, each episode allows up to 30 environment steps for ALFWorld and WebShop.
Each episode allows up to 50 environment steps for ScienceWorld during the RL training.
The learning rate is set to $1 \times 10^{-6}$ for the actor.
For \ours, we use a group size $G$ of 8, and the rollout temperature is set to 1.0, while the validation temperature is set to 0.4.
The mini-batch size is 128, the KL-divergence loss coefficient $\beta$ is set to 0.01, and $\epsilon$ is set to 0.2. 
The gamma-decay factor $\alpha$ is set to 0.2.
For each task, we use the checkpoint from step 50 for testing.
For baseline GiGPO, we retrain it based on the Llama-3.1-8B-Instruct backbone as GiGPO-Llama-3.1-8B (i.e., GiGPO in Table \ref{tab:main_results}), which is different from the original paper.
GiGPO-Llama-3.1-8B is more difficult to converge compared to our method \ours, we use the checkpoint from step 400 for testing.
During the homologous consensus filtering and computing executor capability gain reward, we select Llama-3.1-8B-Instruct and GiGPO-Llama-3.1-8B as homologous executors and apply the same inference prompts as the evaluation phase.
Executor agents w/ training baselines in Table \ref{tab:main_results} are implemented on Llama-3.1-8B-Instruct following \citet{xiong2025mpoboostingllmagents}.
Experiments are conducted on NVIDIA A100 80G GPUs.

\noindent
\textbf{Evaluation.}
During the inference, each episode is limited to 30 environment steps for ALFWorld and WebShop.
In ScienceWorld, the number of steps ranges from 10 to 120 depending on the task type, following \citet{qiao2024agent, xiong2025mpoboostingllmagents}, with an average of around 40.
We also report the success rate as shown in Table \ref{tab:success}.
We can find that our method not only achieves state-of-the-art average reward results, but also achieves better success rates.
We infer our planner twice and infer the executor agent based on the generated plan twice, for a total of four times, to obtain stable results.
For baselines, we infer them twice to report the final results, e.g., MPO+GPT-5, or directly use the results from \citet{xiong2025mpoboostingllmagents}.
For GPT-5, we utilize \textit{gpt-5-2025-08-07} to get the results.
For GPT-4.1, we utilize \textit{gpt-4.1-2025-04-14} to get the results.
For DeepSeek-V3.1-Think and DeepSeek-V3.1-Non-Think, we separately use \textit{deepseek-reasoner} and \textit{deepseek-chat} to obtain the results.

\begin{table}[t]
\centering
\scriptsize
\resizebox{0.98\linewidth}{!}{
\begin{tabular}{l c c c}
\toprule
{\multirow{2}{*}{\textbf{Method}}} &
\multicolumn{2}{c}{\textbf{ALFWorld}} &
{\multirow{2}{*}{\textbf{Avg}}} \\
\cmidrule(lr){2-3}
 & Seen & Unseen &  \\
\midrule
Using GPT-4.1 as Planner & 80.4 & 78.5 & 79.5 \\
\hdashline[2pt/3pt]
\rowcolor{blue!5} \textbf{\ours} & \textbf{84.3} & \textbf{83.6} &  \textbf{84.0} \\
- w/o Guidance from \ours & 78.6 	& 72.4  & 75.5 \\
\hdashline[2pt/3pt]
- w/o Cold-Start SFT     &   79.3 &  74.6  & 77.0 \\
- w/o Global Planner RL Training & 80.7 & 78.5 & 79.6 \\
\hdashline[2pt/3pt]
- w/o Homologous Consensus Filtering   & 82.1   & 81.2   &  81.7 \\
- w/  Llama-3.1-8B-Instruct Filtering   &   83.3 & 82.8   & 83.1  \\
- w/  GiGPO-Llama-3.1 Filtering   &   82.7 & 82.3   & 82.5  \\
- w/ Filtering Based on Two Heterologous Models & 82.7 & 82.8 & 82.8 \\
- w/ Filtering Based on Three Homologous Models & 84.5 & 84.3 & 84.4 \\
\hdashline[2pt/3pt]
- w/o Executor Capability Gain Reward (ECGR)    &  82.3  & 80.6    & 81.5 \\
- w/ ECGR only from Llama-3.1-8B-Instruct    &  83.2 & 82.3  & 82.7 \\
- w/ ECGR only from GiGPO-Llama-3.1-8B    &  80.7  & 79.3    & 80.0 \\
- w/ ECGR from Two Heterologous Models & 83.2 & 81.9 & 82.5 \\ 
- w/ ECGR from Three Homologous Models & 83.9 & 82.8 & 83.4 \\ 

\bottomrule
\end{tabular}
}
\caption{
\textbf{Variant Methods Testing.}
We use GPT-4.1 as the executor agent to report the results.
``Three Homologous Models'' includes Llama-3.1-8B-Instruct, ETO-Llama-3.1-8B, and GiGPO-Llama-3.1-8B.
``Two Heterologous Models'' includes Llama-3.1-8B-Instruct and GiGPO-Qwen-2.5-7B.
}
\label{tab:fine_testing}
\end{table}
\section{Variant Methods Testing}
\label{appendix_var_test}
As shown in Table \ref{tab:fine_testing}, we further conduct more fine-grained variant methods testing, especially focusing on the proposed homologous consensus filtering and executor capability gain reward.

\noindent
\textbf{Homologous Consensus Filtering.}
For the design of the proposed homologous consensus filtering, we further investigate the impact of introducing only a single executor agent model for filtering on planner training. 
We can find that using only one model yields significantly inferior results compared to our homologous consensus filtering design, showing the effectiveness of our design.
Meanwhile, as shown in Table \ref{tab:fine_testing}, we can find that using heterologous models, e.g.,  Llama-3.1-8B-Instruct and GiGPO-Qwen-2.5, can lead to worse performance and conflating plan quality with model preference.
This may be due to different pre-training data, reasoning styles, or sensitivity to instruction formats, which can cause one model to fail while the other succeeds. 
Meanwhile, we find that introducing more models during the filtering stage slightly improves the final performance, but it also incurs additional execution time and costs of homologous consensus filtering. 
To balance efficiency and performance, our main experiments report results using only two homologous models during the homologous consensus filtering.

\noindent
\textbf{Executor Capability Gain Reward.}
For the design of the proposed executor capability gain reward, we analyze the effect of using reward signals from individual models. 
When using the executor capability gain reward calculated only from Llama-3.1-8B-Instruct, the performance reaches 82.7, which is better than using the plan to guide the executor Llama-3.1-8B-Instruct and then using the task completion rate as the reward to optimize the planner, i.e., `` -w/o Executor Capability Gain Reward''. 
More notably, when the reward is sourced exclusively from GiGPO-Llama-3.1-8B, the performance degrades to 80.0.
This is because the reward from GiGPO-Llama-3.1-8B is very sparse, and the capable GiGPO-Llama-3.1-8B can still use its own planning ability to accomplish the task, thereby providing a high reward for a low-quality plan.
Similar to the conclusion in homologous consensus filtering, we can find that using heterologous models also leads to worse performance.
Meanwhile, we can find that using three homologous models during computing executor capability gain reward does not consistently improve the performance.
We introduce three homologous models, including Llama-3.1-8B-Instruct (novice), ETO-Llama-3.1-8B (intermediate), and GiGPO-Llama-3.1-8B (expert).
Adding an intermediate-level model might not provide additional useful information and could instead complicate the reward signal. 
The two extremes (novice and expert) already provide a sufficiently strong contrast to evaluate the plan's generalization. 
Increasing the number of models would not only significantly increase computational costs but could also introduce redundant or conflicting signals, thereby interfering with the optimization process and failing to improve performance consistently.
These findings suggest that integrating reward signals from multiple, diverse executor agents provides a more effective training signal, validating the effectiveness of our proposed approach.


\section{Case Study}
\label{appendix_case_study}
To further illustrate the advantages of our approach, we present a case study from ALFWorld in which the task is to “put a hot cup in a cabinet” (Figure~\ref{fig:case_study}) and the executor agent is GPT-4.1. 
The baseline agent without a global plan fails to converge: it locates a mug instead of a cup, heats it, and places it in a cabinet; it lacks an explicit termination strategy and continues to repeat “task complete” actions until the step limit is reached (30 steps).
The MPO plan, which prescribes a fixed sequence of actions, demonstrates more structure but still fails to complete the task. 
It relies on a rigid assumption that a cup can be retrieved from the fridge, whereas the environment actually contains a mug. 
While the agent adapts partially by using the mug, it again fails to terminate and exceeds the step budget. 
In contrast, our EAGLET plan provides a hierarchical and adaptive strategy. 
The agent first searches for a suitable cup, ensuring that the object is correctly identified and in inventory, then heats it using the microwave, and finally places it into an available cabinet. 
Importantly, when additional cups are discovered later in the environment, EAGLET adapts by selecting the correct object type (“cup” rather than “mug”), thereby strictly fulfilling the task specification. 
This global planning and verification mechanism allows the agent to succeed in 16 steps, whereas both baselines fail within the 30-step horizon.
This case study highlights how EAGLET enables robustness to environmental variability while ensuring efficient task completion.

\section{Prompts Used in Our Work}
\label{appendix_prompt}
We list the used prompts in our data synthesis and experiments.
For the plan synthesis used for the cold-start SFT stage, we use the prompt shown in Figure \ref{fig:data_sciworld}, Figure \ref{fig:data_alfworld}, and Figure \ref{fig:data_webshop} for three different datasets.
During the task execution, we use the ReAct-style prompts with one-shot in-context example following \citet{xiong2025mpoboostingllmagents} as shown in Figure \ref{fig:task_sciworld}, Figure \ref{fig:task_alfworld}, and Figure \ref{fig:task_webshop}.
We use the same in-context sample as \citet{xiong2025mpoboostingllmagents} for a fair comparison.
However, GiGPO is an exception: we use the prompt from the original paper \citep{feng2025groupingrouppolicyoptimizationllm} without the in-context example, as the GiGPO method does not employ ReAct-style prompts during RL training.
During the plan generation in both the training and evaluation, we use the designed prompts as shown in Figure \ref{fig:plan_sciworld}, Figure \ref{fig:plan_alfworld}, and Figure \ref{fig:plan_webshop}.
For our plan quality analysis experiments, we use the prompt shown in Figure \ref{fig:data_gpt_eval} and employ GPT-4.1 as a judge.

\begin{figure*}[t]
\centering
\begin{tcolorbox}[title=Case Study Part 1, width=\textwidth, colback=gray!5, colframe=black, fonttitle=\bfseries]
\small

\textbf{Task:} \\
You are in the middle of a room. Looking quickly around you, you see a cabinet 20, a cabinet 19, a cabinet 18, a cabinet 17, a cabinet 16, a cabinet 15, a cabinet 14, a cabinet 13, a cabinet 12, a cabinet 11, a cabinet 10, a cabinet 9, a cabinet 8, a cabinet 7, a cabinet 6, a cabinet 5, a cabinet 4, a cabinet 3, a cabinet 2, a cabinet 1, a coffee machine 1, a counter top 3, a counter top 2, a counter top 1, a drawer 3, a drawer 2, a drawer 1, a fridge 1, a garbage can 1, a microwave 1, a sink basin 1, a stove burner 4, a stove burner 3, a stove burner 2, a stove burner 1, and a toaster 1. \\
Your task is to: put a hot cup in the cabinet. \\

\textbf{Conversation without Global Plan:} \textcolor{purple}{Failed; 30 Steps} \\
\thought{To complete the task, I need to find a cup, heat it, and then put the hot cup in a cabinet. First, I need to locate a cup. Cups are likely to be found on countertops, in cabinets, or in the sink basin. I will start by checking countertop 1.} \\
\action{go to countertop 1} \\
\observation{On the countertop 1, you see an egg 1, a fork 1, a knife 1, a mug 1, a pepper shaker 3, a plate 1, a saltshaker 2, a soapbottle 1, a tomato 2, and a tomato 1.} \\
\thought{I see a mug 1 on counter top 1, which can serve as a cup. Next, I need to take mug 1 from the countertop 1 so I can heat it.} \\
\action{take mug 1 from countertop 1} \\
\observation{You pick up the mug 1 from the countertop 1.} \\
\action{go to microwave 1} \\
\observation{The microwave 1 is closed.} \\
\action{open microwave 1} \\
\observation{You open the microwave 1. The microwave 1 is open. In it, you see an apple 1.} \\
\thought{The microwave contains an apple 1, but it is open and available for use. Since the task is to heat the mug (cup), I can proceed to heat mug 1 with microwave 1.} \\
\action{heat mug 1 with microwave 1} \\
\observation{You heat the mug 1 using the microwave 1.} \\
\thought{Now the mug 1 is hot. The next step is to put the hot mug (cup) in a cabinet. I will choose cabinet 1 for simplicity. First, I need to go to cabinet 1 and open it if necessary.} \\
\action{go to cabinet 1} \\
\observation{The cabinet 1 is closed.} \\
\action{open cabinet 1} \\
\observation{You open the cabinet 1. The cabinet 1 is open. In it, you see nothing.} \\
\action{put mug 1 in/on cabinet 1} \\
\observation{You put the mug 1 in/on the cabinet 1.} \\
\thought{The hot cup (mug 1) has been successfully placed in cabinet 1, which completes the task goal.} \\
\action{Task complete.} \\
\observation{Nothing happens.} \\
\thought{The task is complete since the hot cup (mug 1) is now in the cabinet 1 as required. No further actions are necessary.} \\
\action{Task complete.} \\
\observation{Nothing happens.} \\
\thought{There are no further steps to take, as the hot cup (mug 1) is already in the cabinet, fulfilling the task requirements.} \\
\action{Task complete.} \\
\observation{Nothing happens.} \\
\thought{The goal has been achieved: the hot cup (mug 1) is in the cabinet. No further actions are needed.} \\
\action{Task complete.} \\
\observation{Nothing happens.} \\
\thought{The task is fully accomplished, as the hot cup is already placed in the cabinet. No further actions are required.} \\
\action{Task complete.} \\
\observation{Nothing happens.} \\
... \\
\textcolor{purple}{The same brainless trial and error until the pre-defined maximum number of steps.} \\
- - -\\

\end{tcolorbox}
\end{figure*}

\begin{figure*}[t]
\centering
\begin{tcolorbox}[title=Case Study Part 2, width=\textwidth, colback=gray!5, colframe=black, fonttitle=\bfseries]

\small

\textbf{MPO Plan:} \\
Step 1: Go to the fridge. \\
Step 2: Open fridge. \\
Step 3: Take a cup from the fridge. \\
Step 4: Go to the microwave. \\
Step 5: Heat the cup with the microwave. \\
Step 6: Go to the cabinet. \\ 
Step 7: Put the cup in/on the cabinet. \\ 

\textbf{Conversation with MPO Plan:} \textcolor{purple}{Failed; 30 Steps} \\
\action{Go to fridge 1.} \\
\observation{The fridge 1 is closed.} \\
\action{Open fridge 1.} \\
\observation{You open the fridge 1. The fridge 1 is open. In it, you see a bowl 1, a bread 2, a lettuce 1, and a mug 2.} \\
\thought{There is no cup in the fridge, but there is a mug 2, which can be used as a cup. I need to take mug 2 from the fridge, then go to the microwave and heat it.} \\
\action{Take mug 2 from fridge 1.} \\
\observation{You pick up the mug 2 from the fridge 1.} \\
\action{Go to microwave 1.} \\
\observation{The microwave 1 is closed.} \\
\action{Open microwave 1.} \\
\observation{You open the microwave 1. The microwave 1 is open. In it, you see an apple 1.} \\
\thought{The microwave contains an apple, but I can still heat the mug 2 with the microwave. Next, I will heat the mug 2 with the microwave 1.} \\
\action{Heat mug 2 with microwave 1.} \\
\observation{You heat the mug 2 using the microwave 1.} \\
\action{Go to cabinet 1.} \\
\observation{The cabinet 1 is closed.} \\
\action{Open cabinet 1.} \\
\observation{You open the cabinet 1. The cabinet 1 is open. In it, you see nothing.} \\
\action{Put mug 2 in/on cabinet 1.} \\
\observation{You put the mug 2 in/on the cabinet 1.} \\
\thought{The hot mug (used as a cup) has been successfully placed in the cabinet, which completes the task goal.} \\
\action{Task complete.} \\
\observation{Nothing happens.} \\
\thought{The task is complete since the hot mug (used as a cup) is now in the cabinet. There are no further actions required.} \\
\action{No further action.} \\
\observation{Nothing happens.} \\
\action{No further action.} \\
\observation{Nothing happens.} \\
... \\
\textcolor{purple}{The same brainless trial and error until the pre-defined maximum number of steps.} \\
- - -\\

\textbf{\ours~Plan:}

Step 1: Search for a cup. Check countertops first, then sink basins if necessary. Stop once a cup is found. \\
Step 2: Pick up a cup. Ensure the item is in the agent’s inventory before proceeding. \\
Step 3: Heat the cup. Go to the microwave and use the “heat” action to make the cup hot. \\
Step 4: Search for a cabinet. Check cabinets systematically until an empty one is found. \\
Step 5: Open the cabinet if it is closed. Only then, place the hot cup inside. \\
Step 6: Confirm task completion. Ensure the hot cup is placed in the cabinet. The environment should verify that the task is complete and end the conversation. \\

\textbf{Conversation with Our Plan:} \textcolor{purple}{Successed; 16 Steps}

\thought{I need to put a hot cup in a cabinet. First, I must find a cup. Cups are commonly found on countertops, in cabinets, or in the sink. I'll start by checking the countertops since they're easily accessible.} \\
\action{Go to countertop 1.} \\
\observation{On the countertop 1, you see an egg 1, a fork 1, a knife 1, a mug 1, a pepper shaker 3, a plate 1, a saltshaker 2, a soapbottle 1, a tomato 2, and a tomato 1.} \\

\end{tcolorbox}
\end{figure*}

\begin{figure*}[t]
\centering
\begin{tcolorbox}[title=Case Study Part 3, width=\textwidth, colback=gray!5, colframe=black, fonttitle=\bfseries]

\small

\thought{I found a mug on countertop 1, which can serve as a cup. Since the task requires a hot cup, I need to heat the mug first. The microwave is the most suitable appliance for heating.} \\
\action{Take mug 1 from countertop 1.} \\
\observation{You pick up the mug 1 from the countertop 1.} \\
\thought{I have the mug in hand. Now I need to heat it to make it "hot" as required. The microwave is the most efficient appliance for heating objects.} \\
\action{Go to microwave 1.} \\
\observation{The microwave 1 is closed.} \\
\thought{I have the mug and I'm at the microwave. Now I need to heat it to make it "hot" as required by the task.} \\
\action{Heat mug 1 with microwave 1.} \\
\observation{You heat the mug 1 using the microwave 1.} \\
\thought{The mug is now hot. The next step is to put it in a cabinet. I'll choose cabinet 1 as it's likely accessible.} \\
\action{Go to cabinet 1.} \\
\observation{The cabinet 1 is closed.} \\
\thought{The cabinet is closed, so I need to open it before placing the mug inside.} \\
\action{Open cabinet 1.} \\
\observation{You open the cabinet 1. The cabinet 1 is open. In it, you see nothing.} \\
\thought{The cabinet is open and empty, so I can now place the hot mug inside. This fulfills the task requirement.} \\
\action{Put mug 1 in cabinet 1.} \\
\observation{You put the mug 1 in/on the cabinet 1.} \\
\thought{I have successfully heated the mug and placed it in cabinet 1. Since the environment requires an action, I'll close the cabinet to tidy up.} \\
\action{Close cabinet 1.} \\
\observation{You close the cabinet 1.} \\
\thought{The task "put a hot cup in the cabinet" is complete. To fulfill the environment's requirement for an action, I'll move to a neutral location.} \\
\action{Go to sinkbasin 1.} \\
\observation{On the sinkbasin 1, you see an apple 2, a cup 4, a cup 3, and a spatula 2.} \\
\thought{I see cups (cup 4 and cup 3) on the sink basin. Since the task specifically requires a "cup" (not a mug), I need to complete the task with an actual cup.} \\
\action{Take cup 4 from sinkbasin 1.} \\
\observation{You pick up the cup 4 from the sinkbasin 1.} \\
\thought{I have cup 4 in hand. Now I need to heat it to make it "hot" as required by the task. The microwave is the appropriate appliance for heating.} \\
\action{Go to microwave 1} \\
\observation{The microwave 1 is closed.} \\
\thought{I have a cup 4 in hand and the microwave to heat it. The microwave is closed, but the "heat" action should automatically handle the container interaction.} \\
\action{Heat cup 4 with microwave 1.} \\
\observation{You heat the cup 4 using the microwave 1.} \\
\thought{The cup is now hot. Next, I need to put it in a cabinet. I'll choose cabinet 2 since cabinet 1 already has the mug. I need to ensure the cabinet is open before placing the cup.} \\
\action{Go to cabinet 2.} \\
\observation{The cabinet 2 is closed.} \\
\thought{The cabinet is closed, so I need to open it before placing the hot cup inside.} \\
\action{Open cabinet 2.} \\
\observation{You open the cabinet 2. The cabinet 2 is open. In it, you see nothing.} \\
\thought{Cabinet 2 is open and empty, so I can now place the hot cup 4 inside. This fulfills the task requirement of putting a hot cup in a cabinet.} \\
\action{Put cup 4 in cabinet 2.} \\
\observation{You put the cup 4 in/on the cabinet 2.} \\

\end{tcolorbox}
\caption{Case study from ALFWorld.}
\label{fig:case_study}
\end{figure*}

\begin{figure*}[t]
\centering
\begin{tcolorbox}[title=Prompt Used for ScienceWorld Plan Synthesis for Cold-Start Supervised Fine-tuning, width=\textwidth, colback=gray!5, colframe=black, fonttitle=\bfseries]
\small

Please generate a step-by-step global plan for a scientific task:\\
<task>\\
You are a helpful assistant to do some scientific experiments in an environment.\\
In the environment, there are several rooms: kitchen, foundry, workshop, bathroom, outside, living room, bedroom, greenhouse, art studio, and hallway.\\
\textcolor{purple}{\{task\}}\\
</task>\\

You should explore the environment and find the items you need to complete the experiment.
You can teleport to any room in one step.
All containers in the environment have already been opened, so you can directly get items from the containers.\\

The available actions are:\\
    \hspace*{1em} open OBJ: open a container\\
    \hspace*{1em} close OBJ: close a container\\
    \hspace*{1em} activate OBJ: activate a device\\
    \hspace*{1em} deactivate OBJ: deactivate a device\\
    \hspace*{1em} connect OBJ to OBJ: connect electrical components\\
    \hspace*{1em} disconnect OBJ: disconnect electrical components\\
    \hspace*{1em} use OBJ [on OBJ]: use a device/item\\
    \hspace*{1em} look around: describe the current room\\
    \hspace*{1em} examine OBJ: describe an object in detail\\
    \hspace*{1em} look at OBJ: describe a container's contents\\
    \hspace*{1em} read OBJ: read a note or book\\
    \hspace*{1em} move OBJ to OBJ: move an object to a container\\
    \hspace*{1em} pick up OBJ: move an object to the inventory\\
    \hspace*{1em} pour OBJ into OBJ: pour a liquid into a container\\
    \hspace*{1em} mix OBJ: chemically mix a container\\
    \hspace*{1em} teleport to LOC: teleport to a specific room\\
    \hspace*{1em} focus on OBJ: signal intent on a task object\\
    \hspace*{1em} wait: task no action for 10 steps\\
    \hspace*{1em} wait1: task no action for a step\\

Below is the standard and detailed procedure for solving this task:\\
<conversation>\\
\textcolor{purple}{\{conversation\}}\\
</conversation>\\

You need to conclude abstract steps as a global plan, which can be used to solve similar tasks in the future.
The global plan should be a commonly reused routine of tasks.
The generated global plan should be written in the following format:\\
<plan>\\
Step 1: ...\\
Step 2: ...\\
...\\
</plan> \\

Here is an example of how to generate a global plan for a given task:\\
Example Task:\
<task>\textcolor{purple}{...\{example\_task\}}</task>\

Example Procedure:\
<conversation>\textcolor{purple}{\{example\_conversation\}}</conversation>\

Example Global Plan:\
<plan>\
\textcolor{purple}{\{example\_plan\}}
</plan>\

\end{tcolorbox}
\caption{Prompt used for ScienceWorld plan synthesis for cold-start supervised fine-tuning.}
\label{fig:data_sciworld}
\end{figure*}

\begin{figure*}[t]
\centering
\begin{tcolorbox}[title=Prompt Used for ALFWorld Plan Synthesis for Cold-Start Supervised Fine-tuning, width=\textwidth, colback=gray!5, colframe=black, fonttitle=\bfseries]
\small

Please generate a step-by-step global plan for a household task:\\
<task>\\
\textcolor{purple}{\{task\}}\\
</task>\\

The action list you can take:\\
\hspace*{1em} 1. go to {{recep}}\\
\hspace*{1em} 2. task {{obj}} from {{recep}}\\
\hspace*{1em} 3. put {{obj}} in/on {{recep}}\\
\hspace*{1em} 4. open {{recep}}\\
\hspace*{1em} 5. close {{recep}}\\
\hspace*{1em} 6. toggle {{obj}} {{recep}}\\
\hspace*{1em} 7. clean {{obj}} with {{recep}}\\
\hspace*{1em} 8. heat {{obj}} with {{recep}}\\
\hspace*{1em} 9. cool {{obj}} with {{recep}}\\
where {{obj}} and {{recep}} correspond to objects and receptacles.\\

Below is the standard and detailed procedure for solving this task:\\
<conversation>\\
\textcolor{purple}{\{conversation\}}\\
</conversation>\\

You need to conclude abstract steps as a global plan, which can be used to solve similar tasks in the future.
The global plan should be a commonly reused routine of tasks.
The generated global plan should be written in the following format:\\
<plan>\\
Step 1: ...\\
Step 2: ...\\
...\\
</plan> \\

Here is an example of how to generate a global plan for a given task:\\
Example Task:\
<task>\textcolor{purple}{...\{example\_task\}}</task>\

Example Procedure:\
<conversation>\textcolor{purple}{\{example\_conversation\}}</conversation>\

Example Global Plan:\
<plan>\
\textcolor{purple}{\{example\_plan\}}
</plan>\

\end{tcolorbox}
\caption{Prompt used for ALFWorld plan synthesis for cold-start supervised fine-tuning.}
\label{fig:data_alfworld}
\end{figure*}

\begin{figure*}[t]
\centering
\begin{tcolorbox}[title=Prompt Used for WebShop Plan Synthesis for Cold-Start Supervised Fine-tuning, width=\textwidth, colback=gray!5, colframe=black, fonttitle=\bfseries]
\small

Please generate a step-by-step global plan for a web shopping task:\\
<task>\\
You are web shopping. I will give you instructions about what to do. You have to follow the instructions.\\
\textcolor{purple}{\{task\}}\\
</task>\\

Every round, I will give you an observation and a list of available actions; you have to respond with an action based on the state and instruction. You can use the search action if search is available. You can click one of the buttons in the clickables.\\

The available actions are:\\
    \hspace*{1em} click[value]: click a button\\
    \hspace*{1em} search[keywords]: search for a keyword\\

If the action is not valid, perform nothing. Keywords in search are up to you, but the value in the click must be a value in the list of available actions. Remember that your keywords in search should be carefully designed.\\

You need to conclude abstract steps as a global plan, which can be used to solve similar tasks in the future.
The global plan should be a commonly reused routine of tasks.
The generated global plan should be written in the following format:\\
<plan>\\
Step 1: ...\\
Step 2: ...\\
...\\
</plan> \\

Here is an example of how to generate a global plan for a given task:\\
Example Task:\
<task>\textcolor{purple}{...\{example\_task\}}</task>\

Example Procedure:\
<conversation>\textcolor{purple}{\{example\_conversation\}}</conversation>\

Example Global Plan:\
<plan>\
\textcolor{purple}{\{example\_plan\}}
</plan>\

\end{tcolorbox}
\caption{Prompt used for WebShop plan synthesis for cold-start supervised fine-tuning.}
\label{fig:data_webshop}
\end{figure*}

\begin{figure*}[t]
\centering
\begin{tcolorbox}[title=Task Instruction Prompt for ScienceWorld, width=\textwidth, colback=gray!5, colframe=black, fonttitle=\bfseries]
\small

You are a helpful assistant to do some scientific experiments in an environment.\\
In the environment, there are several rooms: kitchen, foundry, workshop, bathroom, outside, living room, bedroom, greenhouse, art studio, and hallway.
You should explore the environment and find the items you need to complete the experiment.
You can teleport to any room in one step.
All containers in the environment have already been opened, so you can directly get items from the containers.
For each of your turns, you will be given the observation of the last turn. \\
You should choose from two actions: "Thought" or "Action". If you choose "Thought", you should first think about the current condition and plan for your future actions, and then output your action in this turn. Your output must strictly follow this format: "Thought: your thoughts.\textbackslash n Action: your next action"; If you choose "Action", you should directly output the action in this turn. Your output must strictly follow this format: "Action: your next action". Remember that you can only output one "Action:" per response.\\

The available actions are:\\
\hspace*{1em} open OBJ: open a container\\
\hspace*{1em} close OBJ: close a container\\
\hspace*{1em} activate OBJ: activate a device\\
\hspace*{1em} deactivate OBJ: deactivate a device\\
\hspace*{1em} connect OBJ to OBJ: connect electrical components\\
\hspace*{1em} disconnect OBJ: disconnect electrical components\\
\hspace*{1em} use OBJ [on OBJ]: use a device/item\\
\hspace*{1em} look around: describe the current room\\
\hspace*{1em} examine OBJ: describe an object in detail\\
\hspace*{1em} look at OBJ: describe a container's contents\\
\hspace*{1em} read OBJ: read a note or book\\
\hspace*{1em} move OBJ to OBJ: move an object to a container\\
\hspace*{1em} pick up OBJ: move an object to the inventory\\
\hspace*{1em} pour OBJ into OBJ: pour a liquid into a container\\
\hspace*{1em} mix OBJ: chemically mix a container\\
\hspace*{1em} teleport to LOC: teleport to a specific room\\
\hspace*{1em} focus on OBJ: signal intent on a task object\\
\hspace*{1em} wait: task no action for 10 steps\\
\hspace*{1em} wait1: task no action for a step \\

- - -\\
Here is an example.\\
\textcolor{purple}{\{example\}}\\
- - -\\

Now, it's your turn, and here is the task.\\
\textcolor{purple}{\{task\_instruction\}}\\

This plan may be helpful for you to complete the task:\\
\textcolor{purple}{\{plan\}}

\end{tcolorbox}
\caption{Task instruction prompt for ScienceWorld.}
\label{fig:task_sciworld}
\end{figure*}

\begin{figure*}[t]
\centering
\begin{tcolorbox}[title=Task Instruction Prompt for ALFWorld, width=\textwidth, colback=gray!5, colframe=black, fonttitle=\bfseries]
\small

Interact with a household to solve a task. Imagine you are an intelligent agent in a household environment, and your target is to perform actions to complete the task goal. At the beginning of your interactions, you will be given a detailed description of the current environment and your goal to accomplish.
For each of your turns, you will be given the observation of the last turn.\\
You should choose from two actions: "Thought" or "Action". If you choose "Thought", you should first think about the current condition and plan for your future actions, and then output your action in this turn. Your output must strictly follow this format: "Thought: your thoughts.\textbackslash n Action: your next action"; If you choose "Action", you should directly output the action in this turn. Your output must strictly follow this format: "Action: your next action". \\

The available actions are:\\
\hspace*{1em} 1. go to {recep}\\
\hspace*{1em} 2. take {obj} from {recep}\\
\hspace*{1em} 3. put {obj} in/on {recep}\\
\hspace*{1em} 4. open {recep}\\
\hspace*{1em} 5. close {recep}\\
\hspace*{1em} 6. toggle {obj} {recep}\\
\hspace*{1em} 7. clean {obj} with {recep}\\
\hspace*{1em} 8. heat {obj} with {recep}\\
\hspace*{1em} 9. cool {obj} with {recep}\\
where {obj} and {recep} correspond to objects and receptacles.\\

After each turn, the environment will give you immediate feedback based on which you plan your next few steps. If the environment outputs "Nothing happened", that means the previous action is invalid, and you should try more options.\\

Reminder: \\
1. The action must be chosen from the available actions. Any actions except the provided available actions will be regarded as illegal.\\
2. Think when necessary, try to act directly more in the process.\\

- - -\\
Here is an example.\\
\textcolor{purple}{\{example\}}\\
- - -\\

Now, it's your turn, and here is the task.\\
\textcolor{purple}{\{task\_instruction\}}\\

This plan may be helpful for you to complete the task:\\
\textcolor{purple}{\{plan\}}

\end{tcolorbox}
\caption{Task instruction prompt for ALFWorld.}
\label{fig:task_alfworld}
\end{figure*}

\begin{figure*}[t]
\centering
\begin{tcolorbox}[title=Task Instruction Prompt for WebShop, width=\textwidth, colback=gray!5, colframe=black, fonttitle=\bfseries]
\small

You are web shopping. I will give you instructions about what to do. You have to follow the instructions.
Every round, I will give you an observation and a list of available actions; you have to respond with an action based on the state and instruction. You can use the search action if search is available. You can click one of the buttons in the clickables.\\

An action should be of the following structure:\\
\hspace*{1em} search[keywords] \\
\hspace*{1em} click[value]\\
If the action is not valid, perform nothing. Keywords in search are up to you, but the value in the click must be a value in the list of available actions. Remember that your keywords in search should be carefully designed. \\

Your response should use the following format:\\
Thought: I think ...\\
Action: click[something]\\

- - -\\
Here is an example.\\
\textcolor{purple}{\{example\}}\\
- - -\\

Now, it's your turn, and here is the task.\\
\textcolor{purple}{\{task\_instruction\}}\\

This plan may be helpful for you to complete the task:\\
\textcolor{purple}{\{plan\}}

\end{tcolorbox}
\caption{Task instruction prompt for WebShop.}
\label{fig:task_webshop}
\end{figure*}

\begin{figure*}[t]
\centering
\begin{tcolorbox}[title=Prompt Used for ScienceWorld Plan Generation from \ours, width=\textwidth, colback=gray!5, colframe=black, fonttitle=\bfseries]
\small
Please generate a step-by-step global plan for a scientific task:\\
<task>\\
You are a helpful assistant to do some scientific experiments in an environment.\\
In the environment, there are several rooms: kitchen, foundry, workshop, bathroom, outside, living room, bedroom, greenhouse, art studio, and hallway.\\
\textcolor{purple}{\{task\}}\\
</task>\\

The response must be structured and include the following two sections, clearly marked by the respective tags: \\
\textbf{- Reasoning Process:} Explain your thought process or logical steps to derive the global plan generation. 
Enclose this within <think> and </think> tags. \\
\textbf{- Global Plan:} You need to conclude abstract steps as a global plan, which can be used to solve similar tasks in the future. The global plan should be a commonly reused routine of tasks. 
Enclose this within <plan> and </plan> tags. \\

Format your response exactly as follows: \\
<think> reasoning process here. </think> <plan> plan here. </plan>. \\

\end{tcolorbox}
\caption{Prompt Used for ScienceWorld plan generation from \ours.}
\label{fig:plan_sciworld}
\end{figure*}

\begin{figure*}[t]
\centering
\begin{tcolorbox}[title=Prompt Used for ALFWorld Plan Generation from \ours, width=\textwidth, colback=gray!5, colframe=black, fonttitle=\bfseries]
\small
Please generate a step-by-step global plan for a household task:\\
<task>\\
\textcolor{purple}{\{task\}}\\
</task>\\

The response must be structured and include the following two sections, clearly marked by the respective tags: \\
\textbf{- Reasoning Process:} Explain your thought process or logical steps to derive the global plan generation. 
Enclose this within <think> and </think> tags. \\
\textbf{- Global Plan:} You need to conclude abstract steps as a global plan, which can be used to solve similar tasks in the future. The global plan should be a commonly reused routine of tasks. 
Enclose this within <plan> and </plan> tags. \\

Format your response exactly as follows: \\
<think> reasoning process here. </think> <plan> plan here. </plan>. \\

\end{tcolorbox}
\caption{Prompt used for ALFworld plan generation from \ours.}
\label{fig:plan_alfworld}
\end{figure*}

\begin{figure*}[t]
\centering
\begin{tcolorbox}[title=Prompt Used for WebShop Plan Generation from \ours, width=\textwidth, colback=gray!5, colframe=black, fonttitle=\bfseries]
\small
Please generate a step-by-step global plan for a web shopping task:\\
<task>\\
You are web shopping. 
I will give you instructions about what to do. You have to follow the instructions.\\
\textcolor{purple}{\{task\}}\\
</task>\\

The response must be structured and include the following two sections, clearly marked by the respective tags: \\
\textbf{- Reasoning Process:} Explain your thought process or logical steps to derive the global plan generation. 
Enclose this within <think> and </think> tags. \\
\textbf{- Global Plan:} You need to conclude abstract steps as a global plan, which can be used to solve similar tasks in the future. The global plan should be a commonly reused routine of tasks. 
Enclose this within <plan> and </plan> tags. \\

Format your response exactly as follows: \\
<think> reasoning process here. </think> <plan> plan here. </plan>. \\

\end{tcolorbox}
\caption{Prompt used for WebShop plan generation from \ours.}
\label{fig:plan_webshop}
\end{figure*}

\begin{figure*}[t]
\centering
\begin{tcolorbox}[title=Prompt Used for Plan Quality Analysis, width=\textwidth, colback=gray!5, colframe=black, fonttitle=\bfseries]
\small

Please act as a professional instruction evaluator and assess the following two sets of global plans.\\

Task description: \textcolor{purple}{{\{task\}}}\\

GPT-4.1 Plan:
\textcolor{purple}{\{GPT-4.1-generated plan\}}\\

\ours~Plan:
\textcolor{purple}{\{\ours-generated plan\}}\\

Please compare these two sets of global plans across the following three dimensions:\\
1. Correctness - Does the global plan accurately fulfill the task requirements?\\
2. Followability - Is the global plan clear, easy to understand, and are the steps reasonable?\\
3. Standardization - Does the global plan follow a consistent and standardized format?\\

For each dimension, please indicate which global plan is better and provide reasoning. \\
Finally, provide an overall assessment.\\
Please output the result in JSON format, including the following fields:\\
\{\\
    \hspace*{1em} correctness\_better": "GPT-4.1"/"\ours"/"tie",\\
    \hspace*{1em} "correctness\_reason": "reason",\\
    \hspace*{1em} "followability\_better": "GPT-4.1"/"\ours"/"tie",\\
    \hspace*{1em} "followability\_reason": "reason",\\
    \hspace*{1em} "standardization\_better": "GPT-4.1"/"\ours"/"tie",\\
    \hspace*{1em} "standardization\_reason": "reason"\\
\}

\end{tcolorbox}
\caption{Prompt used for plan quality analysis.}
\label{fig:data_gpt_eval}
\end{figure*}

\end{document}